\pdfoutput=1
\documentclass[journal, 10pt, twocolumn]{IEEEtran}

\usepackage[switch]{lineno}             
\usepackage{cite}                       
\usepackage{amsmath,amssymb,amsfonts}   
\usepackage{graphicx}                   
\usepackage[utf8]{inputenc}             
\usepackage{tabularx}                   
\usepackage{algorithm}
\usepackage{algpseudocode}
\usepackage{mathtools}                  
\usepackage{booktabs}                   
\usepackage{colortbl}                   
\usepackage{microtype}                  
\usepackage{siunitx}                    
\usepackage{xcolor}                     
\usepackage{makecell}
\usepackage[caption=false,font=footnotesize]{subfig}
\usepackage{xurl}
\usepackage[hidelinks]{hyperref}
\usepackage{cleveref}                   
\crefname{figure}{Fig.}{Figs.}
\crefname{table}{Table}{Tables}
\crefname{algorithm}{Algorithm}{Algorithms}
\usepackage{balance}

\newif\ifcoverpage
\coverpagefalse

\ifcoverpage
\else
\usepackage{eso-pic}
\AddToShipoutPictureBG*{%
  \AtPageUpperLeft{%
    \hspace*{0.5\paperwidth}%
    \raisebox{-0.85cm}{%
      \makebox[0pt][c]{%
        \footnotesize\sffamily\color{black!70}%
        This work has been submitted to the IEEE for possible publication.%
      }%
    }%
    \raisebox{-1.15cm}{%
      \makebox[0pt][c]{%
        \footnotesize\sffamily\color{black!70}%
        Copyright may be transferred without notice, after which this version may no longer be accessible.%
      }%
    }%
  }%
}
\fi

\begin{document}
\bstctlcite{BSTcontrol}

\title{Personalized Affective Perception via Continual Learning in Human-Agent Negotiation}

\author{Mehmet~Onur~Keskin,
Berk~Buzcu,
Nikhil~Churamani,
Reyhan~Aydo\u{g}an,
Hatice~Gunes,
and~Catholijn~Jonker
\thanks{M. O. Keskin and R. Aydo\u{g}an are with the Dept. of Computer Science and Artificial Intelligence \& Data Engineering, \"{O}zye\u{g}in University, Istanbul, T\"{u}rkiye. M. O. Keskin undertook part of this research work while he was a visiting MSc. student at the AFAR Lab, University of Cambridge, UK, and the Interactive Intelligence Group at TU Delft, The Netherlands.}\thanks{B. Buzcu is with the Dept. of Computer Science, HES-SO Valais-Wallis, Sierre, Switzerland.}\thanks{N. Churamani and H. Gunes are with the Dept. of Computer Science and Technology, University of Cambridge, Cambridge, UK.}\thanks{C. M. Jonker and R. Aydo\u{g}an are with Interactive Intelligence, Delft University of Technology, Delft, Netherlands.\\
\textbf{Contributions:} Conceptualization: MOK, RA, HG, CMJ. Methodology: MOK, NC, RA, HG, CMJ. Software: MOK, BB, NC. Formal analysis: MOK. Investigation \& Data Curation: MOK, BB. Writing - Original Draft: MOK. Writing - Review \& Editing: MOK, RA, HG, CMJ. Visualization: MOK. Supervision: RA, HG, CMJ. Project administration: RA, HG. Funding acquisition: RA, HG.
}}

\maketitle

\begin{abstract}
Self-interested intelligent agents that negotiate with humans should interpret nonverbal expressions and adapt their behavior accordingly. However, generic facial-affect models degrade across individuals and contexts, creating a generalization gap.
We introduce an affect-adaptive negotiation strategy that incorporates Continual Learning with Imagination for Facial Expression Recognition (CLIFER) to personalize valence-arousal perception in real time. Differential feedback converts changes in valence-arousal direction and intensity into adjustments of the agent's target utility. An asynchronous architecture updates CLIFER in the background during live interaction. In a counterbalanced within-subject study in which 66 participants negotiated with a humanoid robot, the CL-enabled agent achieved significantly higher individual utility and elicited higher participant concession rates than the FaceChannel baseline. The CL condition yielded significantly higher mean arousal estimates, alongside a broader output range. Participant utility and Nash distance did not change significantly between conditions. These outcomes support the practical value of coupling continual affective personalization with differential feedback for real-time negotiation.

\end{abstract}

\section{Introduction}

\label{sec:introduction}

Negotiating resources, tasks, and conflicting interests is a fundamental capability for socially intelligent agents collaborating with humans \cite{kraus2001, fatima14}. Traditionally, Automated Negotiation has been modeled with Game Theory and AI, viewing agents as rational utility maximizers that exchange offers based on fixed mathematical functions, heuristics, or reinforcement learning policies \cite{rubinstein1982, faratin98}.
Recent agents demonstrate sophisticated strategic reasoning in human-agent negotiation: from the tool-integrated ASTRA agent that performs dynamic offer optimization \cite{kwon2025} to competitive bargaining models like ChargingBoul \cite{shymanski2025} and systems that analyze strategic trade-offs in multi-agent environments \cite{qian2025}. These approaches advance strategic decision-making, while situated human interaction also requires attention to individual preferences and affective cues \cite{broekens2010,demelo2014}. Balancing utility with human social acceptance is still a challenge \cite{Lin2010,baarslag2016}. As embodied agents transition from industrial environments to unstructured social roles (e.g., mediators, tutors, and caretakers), affective intelligence becomes a functional necessity, and expressions serve as social signs of the interaction partner's affective state. 
For example, a tense or serious expression may signal resistance before a rejection is verbalized. At the same time, a slight or mild smile may indicate mutual agreement \cite{yuasa2007, vanKleef2009}. Recent work has begun to utilize this, with agents using affect for dispute resolution \cite{rakshit2025} and persuasion \cite{wang2024}.

Affect perception is not only perspective-dependent but also highly personal. Models built on static datasets commonly fail to detect the dynamics of a social task in which the observer is an active participant. This failure creates what is known as a generalization gap~\cite{gunes2023}. Gunes et al. point out that high accuracy on static benchmarks does not guarantee functional adaptation in the wild.
Generic models such as FaceChannel~\cite{faceChannel2020} provide a common affect representation, but contextual interpretation remains difficult \cite{gunes2023}; for instance, a concentrated expression need not indicate anger. In a negotiation context, this is critical: if the agent misinterprets concentration as anger, it may alter its concession target in response to an ambiguous signal.
This need for adaptation raises the plasticity-stability dilemma: the agent needs plasticity to learn the current participant's idiosyncratic expressions, while excessive adaptation can lead to catastrophic forgetting of previously learned affective knowledge.

We address this challenge by adapting affect perception to observations from the ongoing negotiation, with generative replay to support retention of general affective representations.
Our agent integrates continual learning (CL) to personalize affective perception. A CL-enabled system retains general knowledge while incrementally updating its parameters using the current participant's facial observations \cite{clifer2020, Churamani2020}. We pursue three objectives: (i) to enhance negotiation efficiency through personalized perception, (ii) to improve human-agent negotiation outcomes, and (iii) to validate continual learning as allowing real-time adaptation for affect understanding. To realize these objectives, we extend the Solver Agent~\cite{solver2021,keskin2025} from categorical affective feedback to differential feedback from continuous valence-arousal estimates. FaceChannel uses a fixed pretrained affect model, while CLIFER combines participant-specific demo training with continued adaptation during negotiation; both share the differential-feedback policy.

We compare these two perception conditions in human-agent negotiation under the following Research Questions (RQs):

\begin{itemize}

\item

\textbf{RQ1 (Affective Feedback):} \textit{Does continual learning produce a behaviorally useful affective signal versus generic models?} We analyze how personalized adaptation changes arousal/valence estimates and whether these representations produce affect signals influencing negotiation behavior.

\item

\textbf{RQ2 (Behavioral Influence):} \textit{How does the agent’s personalized behavior influence human negotiation behavior?} We examine whether an agent that adapts its affective feedback during the interaction induces a behavioral shift. We explore whether changes in human concession rates statistically mediate differences in agent utility.

\item

\textbf{RQ3 (Negotiation Efficiency):} \textit{Does personalized affective perception enable the agent to gain higher individual utility without degrading joint optimality (Nash distance) in fewer negotiation rounds?} We test whether adapting perception improves the agent's return and negotiation speed, and examine the accompanying participant utility and Nash distance. We hypothesize that personalized affective feedback helps the agent adapt its strategy early and reach agreements efficiently.

\item

\textbf{RQ4 (Questionnaire Analysis):} \textit{Do participants perceive the CL-enabled agent as more sophisticated?} We examine participants' ratings of preference modeling, affective awareness, and persuasiveness to assess how personalized affective perception shapes their experience of its negotiation behavior.

\end{itemize}

Our methodological contribution is a differential affective feedback rule that turns angular and radial changes in continually personalized valence-arousal estimates into negotiation target-utility adjustments. We couple this rule to asynchronous CLIFER updates within the Solver Agent and Negotiator framework \cite{clifer2020,solver2021,keskin2025,negotiator2024}, keeping perceptual adaptation active during the exchange. In a 66-participant human-robot study using a shared negotiation policy, the CL-enabled system achieved higher agent utility and elicited higher participant concession rates than the fixed-perception baseline.

\section{Related Work}

\label{sec:related}

Developing socially intelligent agents that negotiate with humans requires combining multiple research areas, including negotiation strategies, affective perception, and personalization.

\subsection{Human-agent negotiation beyond utility maximization}

Automated negotiation research has historically relied on heuristics and game-theoretic models \cite{fatima06, hindriks2011}.
Competitions such as the Automated Negotiating Agents Competition (ANAC) \cite{anac2019} provide testbeds for bidding strategies and opponent models. Recent work has also introduced Large Language Models (LLMs) for offer selection and opponent inference in human-agent negotiations. Kwon et al. \cite{kwon2025} introduced ASTRA for tool-integrated offer optimization. In contrast, Kuru et al. \cite{kuru2025evaluating} evaluated the Theory-of-Mind (ToM) capacities of LLMs by testing their ability to infer opponents' preferences from offer histories. Opponent modeling also guides offer selection \cite{keskin2023conflict}. ChargingBoul \cite{shymanski2025} classifies opponents by their bid patterns and adapts its strategy accordingly. Human-Agent Negotiation (HAN) also requires attention to social signals and the relationship between utility and interpersonal goals \cite{aydogan2022, baarslag2022}. Frameworks such as Negotiator \cite{negotiator2024} and IAGO \cite{IAGO2016} provide reusable environments for studying human-agent negotiation. Vahidov et al. \cite{vahidov2017} examined how concession schedules and task complexity affect agreements. Argument-aware agents incorporate opponents' arguments into modeling and bidding \cite{dogru2025}; robot-clerk research studies how agents acquire sales strategies for a particular venue \cite{muramatsu2025}.

\subsection{Affective feedback as a strategic signal in negotiation}

In this interactive setting, affective feedback guides negotiation decisions \cite{brown2013, demelo2023}. Grounded in the ``Emotions as Social Information'' (EASI) model, the literature shows that emotional displays influence an observer's behavior through both inferential and affective routes \cite{vanKleef2009}. Early works demonstrated that agents expressing anger or happiness could systematically manipulate human concession rates and shift core appraisal mechanisms \cite{demelo2014, aggressiveAgent2015}. Emotion also conveys actionable strategic data. Observing an AI's emotional reactions during pre-play communication allows humans to reverse-appraise the agent's hidden preferences, unlocking highly integrative agreements \cite{sato2023}. Terada et al. \cite{terada2025} recently proposed a Bayesian approach for inferring a partner's hidden Social Value Orientation (SVO) from their emotional reactions to offers.

In volatile dispute-resolution scenarios, tracking affective trajectories proves critical. Unmanaged anger often traps negotiating parties in a destructive loop that leads to a deadlock, whereas expressing compassion can defuse tension and guide the interaction toward an agreement \cite{rakshit2025}. Wang et al. \cite{wang2024} similarly demonstrated that emotion-driven concession curves can optimize multi-objective outcomes. Affective signaling also influences subjective evaluations of the interaction. Nonverbal expressions like smiles, virtual touch \cite{harjunen18}, and agent warmth \cite{prajod19} elevate human compliance, perceived fairness, and the willingness to renegotiate, even when economic outcomes remain constant. Fine-grained affective traits and their timing explain significant variance in subjective negotiation outcomes, including participant satisfaction and partner perception \cite{chawla2021, chawla2024}. Recent embodied negotiation systems integrate facial affect directly into their strategies \cite{keskin2025}.

\subsection{Affect recognition and limitations}

Using emotion strategically requires accurate perception, which traditional affective computing often struggles to provide. This interaction involves both perceiving the participant's expressions and displaying cues that the participant can interpret.
Physical presence can affect responses to social agents \cite{lee2006, LI2015, cakan2023embodiment}, while robot appearance can affect negotiation dynamics \cite{keskin2024}.
Generic emotional displays on a physical agent can also be difficult to interpret. Rendering intense, generic expressions of joy on an android can trigger the uncanny valley effect, impairing a human's ability to understand affect and infer preferences compared to interacting with a screen-based virtual agent \cite{sato2024}.
Cultural differences make this issue more complex.
Strategies effective in one context may fail in another because culturally conditioned responses conflict \cite{kopelman2008, gal2011, haim2012}. This expressive variation motivates personalizing perception \cite{Takayama2009}.

Despite progress in personalization, an important bottleneck for affective agents remains the generalization gap \cite{gunes2023}.
Generic models such as FaceChannel \cite{faceChannel2020} and prior multimodal systems \cite{aly16} provide affect estimates whose interpretation depends on the interaction context.
Past research has emphasized emotional opponent modeling \cite{hoegen2017, Sebastiani2017}, but has generally relied on generic interpretations. A growing consensus in affective science argues that imposing rigid emotion categories (e.g., the ``Six Basic Emotions'') on facial movements is fundamentally flawed \cite{barrett2019}. In real negotiations, a knitted brow or concentrated expression may indicate cognitive effort rather than categorical anger \cite{gunes2023}.
Continuous valence-arousal models represent graded affective changes, but their estimates still require contextual interpretation during interaction \cite{gunes2023}.
For example, a participant's baseline expression can dynamically change as they feel tired or build rapport.
Arousal alone does not distinguish negative from positive affect; a control strategy must also account for valence and changes in the two-dimensional signal. A fixed mapping from affect estimates to actions can therefore miss changes that matter for negotiation.

\subsection{Personalization and continual learning for Affective HRI}

Continual learning offers an approach to adapting perception during interaction and combines incremental adaptation with mechanisms to mitigate Catastrophic Forgetting, the loss of previously learned knowledge when acquiring new information~\cite{clifer2020}. In affective robotics, this balance concerns adapting to the current participant while retaining general facial representations.
Churamani et al. \cite{Churamani2020} established the computational basis for solving this in affective robotics using dual-memory architectures.
Frameworks like CLIFER (Continual Learning with Imagination for Facial Expression Recognition) \cite{clifer2020} operationalize this approach, enabling an embodied system to construct a coherent, personalized affective signal amid the variability of live interaction. Related personalization research outside negotiation reports that fusing generalized baselines with person-specific affective adaptations significantly improves real-time interpretative accuracy \cite{spaulding2019, barros2022}. Latent generative replay reduces the cost of facial-expression learning by generating intermediate feature embeddings instead of images \cite{Stoychev2023LGR}. Our system uses image-level replay with asynchronous updates to keep learning off the interaction path.

Affective signals influence concession behavior and participants' evaluations of negotiation counterparts \cite{chawla2024, rakshit2025}. Prior work in affective robotics has used continual personalization in wellbeing interactions \cite{Churamani2022CL4HRI} and affect-driven policy learning in an extended Ultimatum Game \cite{churamani2022}. More recent personalized facial-expression recognition addresses source-to-person adaptation \cite{zeeshan2026progressive}. We connect online affective personalization to concession control in bilateral, multi-issue negotiation. Our differential-feedback rule converts directional and radial changes in adapting valence-arousal estimates into target-utility adjustments, extending the categorical feedback of the earlier Solver \cite{solver2021,keskin2025}. Comparing fixed and adapting perception under this shared rule tests the operational value of personalization within a live negotiation loop.
\section{Background}

\label{sec:background}

We first define the negotiation setting, outcome metrics, and offer-move categories, then describe the \textit{Solver Agent} policy. Finally, we detail the computational models used for affective perception: \textit{FaceChannel} as our static, generic baseline and \textit{CLIFER} as our continual learning framework for online adaptation with generative replay.

\subsection{Bilateral Multi-Issue Human-Agent Negotiation}

The automated negotiation literature generally models the interaction as a bilateral, multi-issue negotiation between a human ($h$) and an agent ($a$) using the Alternating Offers Protocol (AOP)~\cite{rubinstein1982}, in which participants take turns exchanging offers. The negotiation domain consists of a set of issues $I = \{i_1, \dots, i_m\}$, where each issue $i_j$ ($j=1,\ldots,m$) takes values from a predefined domain $D_{i_j}$. An offer $\omega = (v_1, \dots, v_m)$ is a tuple where each $v_j \in D_{i_j}$ represents the specific value assigned to issue $i_j$. Each participant $x \in \{h, a\}$ has private preferences modeled by a linear-additive utility function. The utility of an offer $\omega$ for participant $x$ is defined as:

\begin{equation}
U_x(\omega) = \sum_{j=1}^{m} w_{j,x} \cdot V_{j,x}(v_j)
\end{equation}

Here, $w_{j,x}$ represents the normalized importance weight of issue $i_j$ (such that $\sum_{j=1}^{m} w_{j,x} = 1$), and $V_{j,x}: D_{i_j} \to [0, 1]$ is the evaluation function mapping the selected value $v_j$ to a normalized score. Negotiation proceeds in rounds until an agreement is reached or the deadline $T_{\max}$ expires.

To evaluate joint-outcome optimality, we use the Nash Distance, defined as the Euclidean distance between the actual agreement $\omega$ and the Nash Bargaining Solution (NBS) on the Pareto frontier. Lower values indicate closer proximity to the NBS.
Formally, assuming a disagreement point of $(0,0)$, the NBS is the optimal outcome $\omega^*$ that maximizes the product of both parties' utilities:

\begin{equation}
\omega^* = \arg\max_{\omega} \big( U_a(\omega) \cdot U_h(\omega) \big)
\end{equation}

The Nash Distance ($D_N$) for a reached agreement $\omega$ is:

\begin{equation}
D_N(\omega) = \sqrt{[ U_a(\omega)-U_a(\omega^*)]^2+[ U_h(\omega)-U_h(\omega^*)]^2}
\label{eq:nash_distance}
\end{equation}

\begin{figure}[ht]
\centerline{\includegraphics[width=\columnwidth]{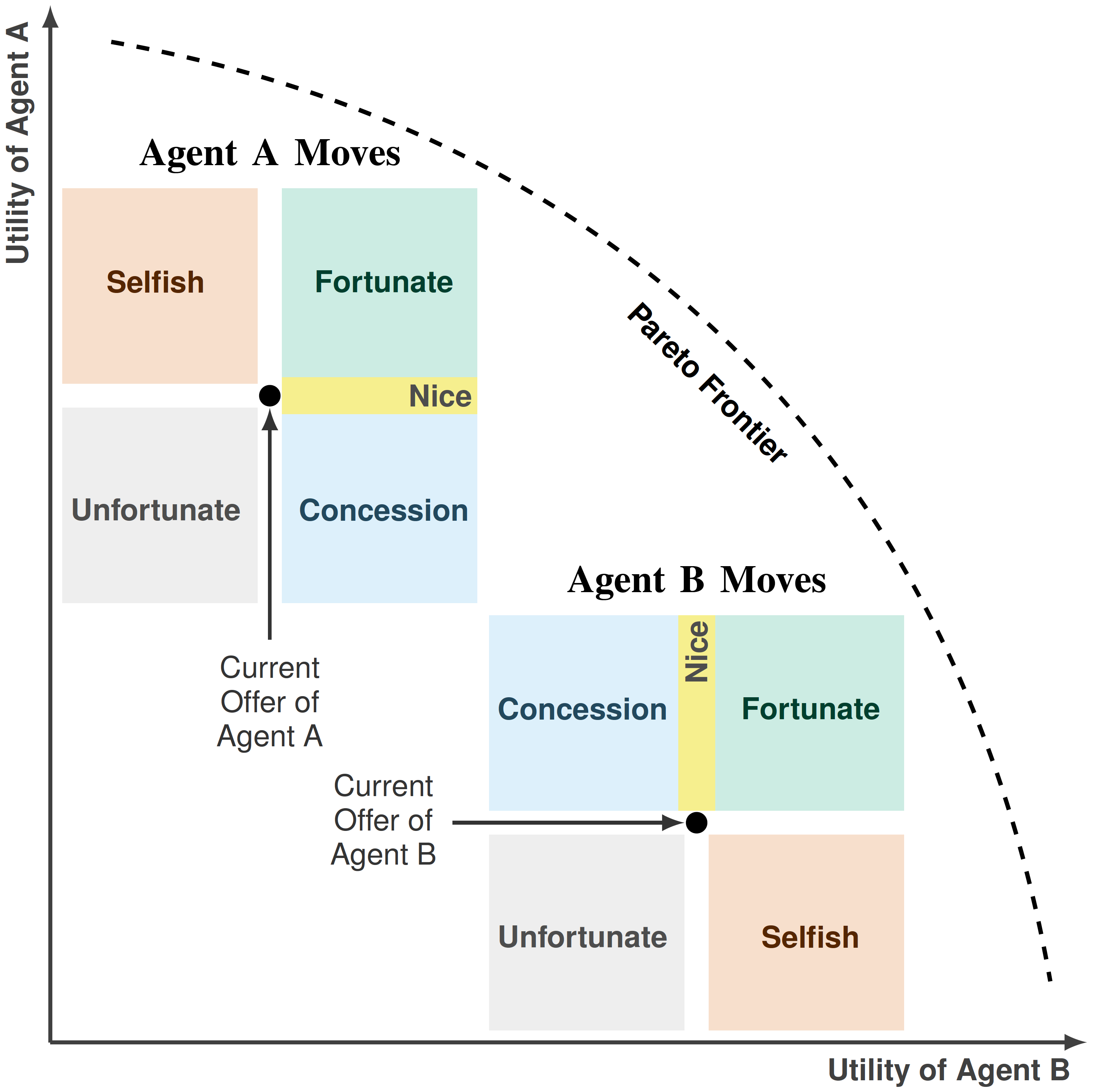}}
\caption{Stepwise Move Classification~\cite{hindriks2011}: Illustrated move types are defined by both parties' utility changes from the same actor's previous offer: Fortunate (both gain), Nice (opponent gains, own unchanged), Concession (own loss, opponent gain), Selfish (own gain, opponent loss), Unfortunate (both lose).}
    \label{fig:stepwise_analysis}
\end{figure}

While automated negotiators traditionally prioritize self-sufficiency, they struggle with interdependence, such as sustaining cooperative social relationships. Effective interdependence requires the agent to respond adaptively to the participant’s verbal and non-verbal feedback.
We conceptualize the agent’s role in terms of self-sufficiency, self-directedness, and interdependence, as Baarslag et al.~\cite{baarslag2022} propose.
To analyze negotiation behaviors beyond final utility scores, we used the stepwise analysis framework proposed by Hindriks et al.~\cite{hindriks2011}. As shown in \cref{fig:stepwise_analysis}, this method classifies offer transitions by the utility difference ($\Delta U$) from the same actor's previous offer. A \textit{Silent} move leaves both parties' utilities unchanged from the same actor's preceding offer.

\subsection{The Solver Agent}

\label{subsec:solver_background}

To compare affective perception within a shared strategy, we use the Solver Agent’s self-directed, deadline-driven tactics and focus on affective feedback.
The Solver Agent’s Target Utility ($TU$) at normalized time $t \in [0,1]$ is defined as a convex combination of a time-dependent tactic ($TU_{Time}$) and a behavior-dependent tactic ($TU_{Be}$):

\begin{equation}
TU(t) = t^2 \cdot TU_{Time}(t) + (1-t^2) \cdot TU_{Be}(t)
\label{eq:hybrid_mix}
\end{equation}

This weighting method lets the agent focus on affective features and the opponent’s offer behavior during early negotiation phases, while gradually shifting toward deadline-driven concession as the interaction concludes.
As the deadline approaches ($t \to 1$), the behavioral component weight $(1-t^2)$ decays.
This ensures that even if the perception component detects high arousal (e.g., excitement or anxiety) in the final rounds, or the participant’s offer behaviors fluctuate, the agent focuses on the deadline-driven concession ($TU_{Time}$).

The underlying Solver strategy \cite{solver2021,keskin2025} defines $TU_{Be}$, which combines the opponent's awareness score ($P_A$), affective feedback ($P_E$), and offer-utility changes. The earlier Solver computes $P_E$ as a weighted sum of emotion-category probabilities averaged across frames~\cite{keskin2025}. Section~IV replaces that calculation with differential feedback from continuous valence-arousal changes.
Additionally, the Solver agent improved the standard Tit-for-Tat strategy by gating the influence of emotion based on the consistency of the opponent’s behavior:

\begin{equation}
TU_{Be} = U_a(\omega_a^{t-1}) + \underbrace{P_A^2 \cdot P_E}_{\text{Affective Feedback}} - \underbrace{(1 - P_A^2)(\mu \cdot \Delta U)}_{\text{Behavioral Feedback}}
\label{eq:solver_behavior}
\end{equation}

\begin{figure*}[t!]
\centering
\includegraphics[width=\linewidth]{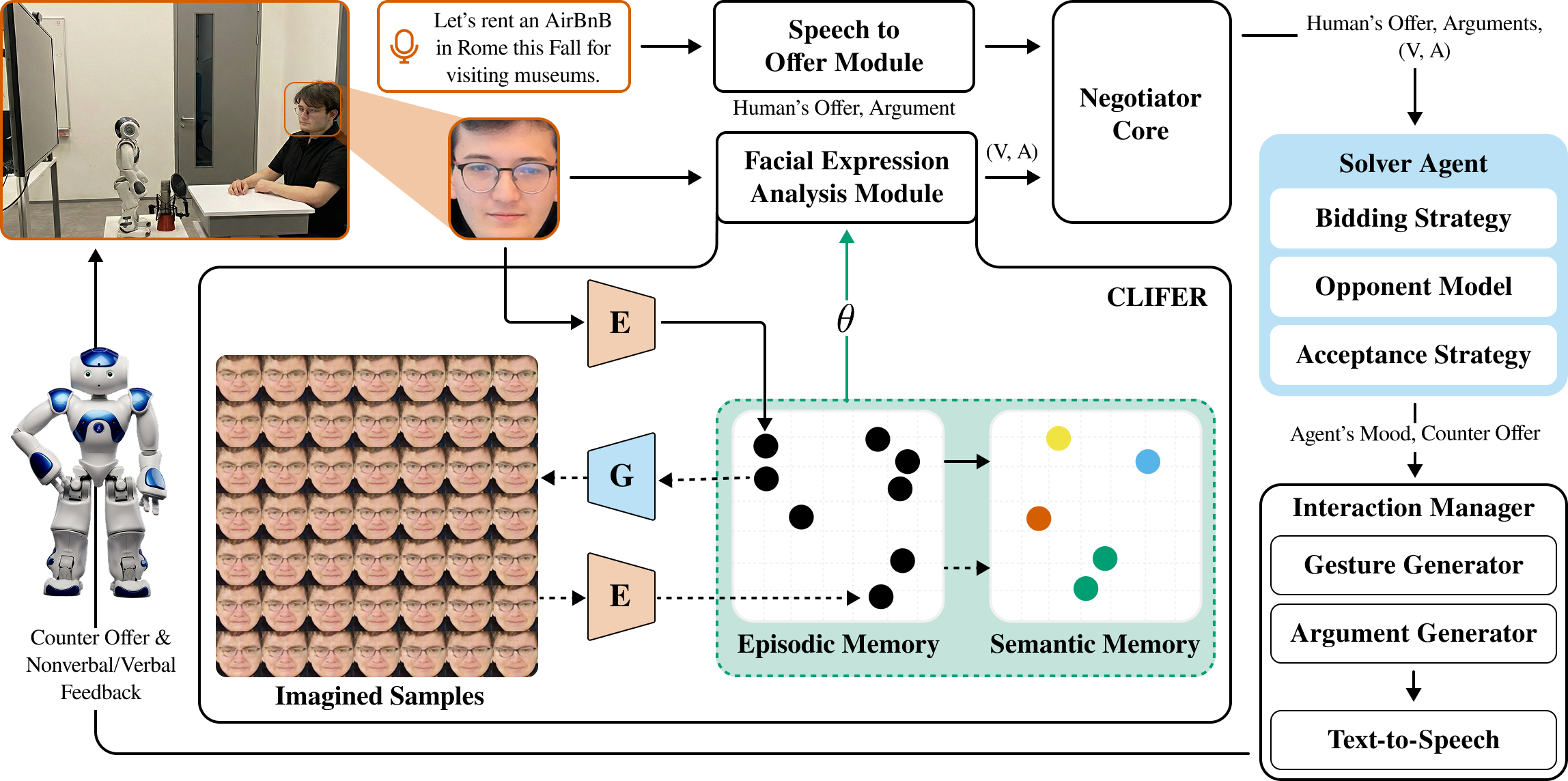}
\caption{System Architecture of the CL-enabled Solver Agent: The robot's audiovisual input is converted to offers (Speech-to-Offer) and valence–arousal estimates (CLIFER: episodic memory adapted to the participant, CAAE providing generative replay). The Solver uses these to compute counter-offers, which the Interaction Manager voices with gestures. $\theta$ denotes the GWR memory parameters; the FaceChannel encoder and regression head remain fixed.}
\label{fig:architecture}
\end{figure*}

Here, $U_a(\omega_{a}^{t-1})$ is the utility of the agent's previous offer, and $\mu \in [0,1]$ is the empathy factor weighting the behavioral feedback. It increases with normalized time as $\mu=P_3(1+t)$, with $P_3=0.5$, following the underlying Solver policy~\cite{keskin2025}. The two dynamic terms function as follows:

\begin{itemize}

\item

\textbf{Affective Feedback ($P_A^2 \cdot P_E$):} Weights the affective feedback ($P_E$) by the opponent's behavioral responsiveness ($P_A$). Positive affective feedback ($P_E > 0$) encourages the agent to maintain its position, while negative feedback ($P_E < 0$) encourages concession to avoid impasse.

\item

\textbf{Behavioral Feedback ($-(1-P_A^2)(\mu\cdot\Delta U)$):} Adjusts target utility in response to changes in the opponent's offers. If the opponent is less responsive to the agent's behavior (i.e., low $P_A$), the agent assigns less weight to the emotional signal and places more weight on the opponent's offer behavior. This Tit-for-Tat term uses $\Delta U$, a weighted sum of recent signed changes in the agent's utility across the opponent's offers. The empathy factor $\mu$ scales this behavioral response. Positive changes lower this term's contribution to the target; negative changes raise it.

\end{itemize}

This control mechanism increases the influence of social signals as behavioral responsiveness ($P_A$) rises.
The awareness score measures the opponent’s response to the agent’s behavior changes~\cite{keskin2025}. The ratio $C_H/C_A$ is clipped to $[0.25,0.75]$, where $C_A$ counts agent move-type changes and $C_H$ counts those followed by an opponent shift. The score is initially $0.5$. Online move classification uses an opponent-utility model; the participant’s true preferences remain private.

\subsection{Generic Affective Baseline: FaceChannel}

To recognize nuanced affective shifts without forcing expressions into rigid categories, our strategy relies on continuous Valence-Arousal (VA) representations. FaceChannel~\cite{faceChannel2020} is a lightweight convolutional neural network pretrained on AffectNet for continuous VA regression. We selected it as our generic baseline because its continuous circumplex output aligns directly with our affective strategy, and its lightweight architecture allows real-time inference within our $<2.0$\,s latency constraint without cloud APIs. It outputs predictions in the range $[-1, 1]$ for both arousal and valence dimensions.

In our within-subjects design, FaceChannel serves as the generic baseline and does not adapt in real time during negotiation. CLIFER starts from a model trained on the participant's demo session (Section~\ref{subsec:demo}) and continues adapting during its own session. The perception results therefore compare outputs from the respective FC and CL sessions, rather than paired predictions on identical video frames. The selected model's output influences the agent's offers and, consequently, the participant's subsequent behavior and facial input.

\subsection{Adaptive Perception: CLIFER with Generative Replay}

Continuous fine-tuning updates pretrained model weights, which can cause catastrophic forgetting. To address the plasticity-stability dilemma in Section \ref{sec:related}, we employ the CLIFER framework \cite{clifer2020}. Its dual-memory architecture has complementary roles for adaptation and retention:

\begin{itemize}

\item

\textbf{Episodic Memory (GDM-E) for Plasticity:} Acting as a dynamic buffer, this module uses unsupervised Hebbian learning via Growing-When-Required (GWR) memory networks to rapidly adapt to the participant's live frames $F_r$ during a negotiation. This supports adaptation to the current participant's observations.

\item \textbf{Semantic Memory (GDM-S) for Stability:} A second GWR memory retains general affective representations. A separate Conditional Adversarial Auto-Encoder (CAAE), pretrained on AffectNet, generates replay images to support retention during adaptation.

\end{itemize}

To mitigate catastrophic forgetting while enabling this plasticity, we use Generative Replay. As illustrated in \cref{fig:architecture}, the system mixes live samples ($F_r$) with ``Imagined'' samples ($F_{gen}$) generated by the CAAE.
Both GDM networks utilize competitive Hebbian learning.
For each input, the system calculates the Best-Matching Unit (BMU) activation.
Neuron creation uses the activation thresholds ($a_E=0.6$ for Episodic, $a_S=0.5$ for Semantic) together with habituation criteria. Otherwise, the system updates the existing BMU and its neighbors.
The GWR memories update their stored Valence-Arousal (VA) associations using pseudo-labels from the frozen FaceChannel model. Its 200-dimensional encoder and regression head remain fixed; adaptation takes place in the memories, without manual affect annotations.
GDM-E aggregates frame-level pseudo-labels through participant-specific clustering. These cluster-level estimates supply adaptive affective input to the negotiation controller.
\section{Proposed Approach: Differential Affective Feedback}

\label{sec-strategy}

In negotiation, the agent must respond to its changing affective input. High arousal may accompany either positive or negative valence, so arousal alone does not determine the response direction. We formalize a \textit{Differential Affective Feedback} ($P_E$) signal that combines changes in direction and intensity. Rotation ($P_{rot}$) captures directional changes in the estimated affective vector, while radial change ($P_{rad}$) captures changes in its magnitude. This feedback rule connects the continuous valence-arousal representation to the strategy’s response \cite{russell1980, vanKleef2009}.

\begin{figure}[ht]
\centerline{\includegraphics[width=\columnwidth]{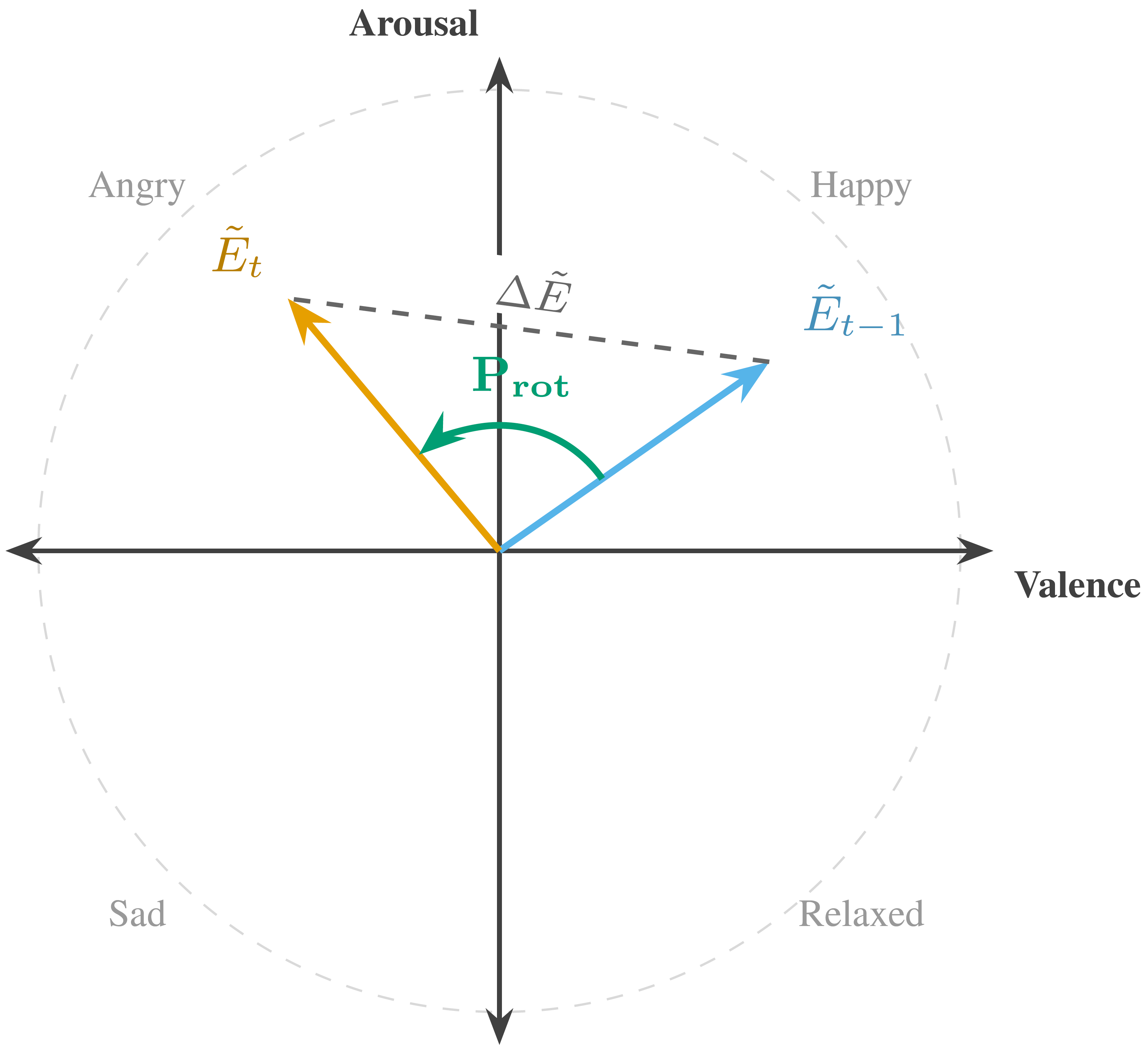}}
\caption{Differential affective feedback between the previous ($\tilde E_{t-1}$, blue) and current ($\tilde E_t$, orange) normalized VA vectors. Equation~\ref{eq:rot} defines angular ($P_{rot}$) and radial ($P_{rad}$) features; the dashed line shows displacement ($\Delta\tilde E$).}
\label{fig:differential_feedback}
\end{figure}

The \textit{Differential Affective Feedback} combines the directional sign with the Euclidean combination of these components:

\begin{equation}
P_E = P_{sign} \times \min\left\{1, \sqrt{P_{rot}^2 + P_{rad}^2}\right\}
\label{eq:PE_final}
\end{equation}

\noindent where $P_{sign}$ sets the direction of the affective target-utility adjustment, while angular and radial changes determine its magnitude. The sign of the rotation is immaterial in Equation~\ref{eq:PE_final}, which uses $P_{rot}^2$. The behavioral term and time-dependent tactic also contribute to the final target utility.
We compute $P_{sign}$ from the changes in normalized valence and arousal, $\Delta_V$ and $\Delta_A$: $P_{sign}=+1$ when $\Delta_V+\Delta_A\geq0$, and $-1$ otherwise. This equal weighting allows either dimension to affect the response; when the changes oppose each other, the change with the larger absolute value determines the sign; ties give $+1$. The circumplex describes affective states, while this control convention acts on changes between states. Before Equation~\ref{eq:PE_final}, we divide $P_{rot}$ by $\pi$ and $P_{rad}$ by $\sqrt{2}$ (ScaleFeedback in Algorithm~\ref{Alg:solver_cl}); the cap bounds $|P_E|$ at one.

From consecutive states $\tilde{E}_{t-1}$ and $\tilde{E}_t$, we compute the unscaled angular change ($P_{rot}$) with the four-quadrant arctangent, and the radial change ($P_{rad}$) from their vector magnitudes:

\begin{equation}
\begin{aligned}P_{rot} &= \operatorname{atan2}(\delta_{det},\delta_{dot}),\\P_{rad} &= \big|\|\tilde{E}_t\|_2-\|\tilde{E}_{t-1}\|_2\big|.\end{aligned}
\label{eq:rot}
\end{equation}

Here, $\delta_{det} = \det(\tilde{E}_{t-1}, \tilde{E}_t)$, $\delta_{dot} = \tilde{E}_{t-1} \cdot \tilde{E}_t$, and $\|\cdot\|_2$ denotes the $L_2$ norm.
This explicit separation mathematically distinguishes directional changes from intensity changes in the estimated affective signal.
To enable fair comparison across participants with different expressive baselines, we first normalize the raw affect values ($E_t$) during their negotiation session as seen in Equation \ref{eq:normalization}:

\begin{equation}
\tilde{E}_t = \frac{E_t - \boldsymbol{\mu}_t}{\boldsymbol{\sigma}_t + \epsilon}
\label{eq:normalization}
\end{equation}

\noindent where $\tilde E_t=(\tilde V_t,\tilde A_t)$, $\boldsymbol{\mu}_t$ and $\boldsymbol{\sigma}_t$ are the running mean and standard deviation vectors of the raw model outputs, and $\epsilon$ is a small positive constant that prevents division by zero.
This adjustable scaling lets the geometric origin $(0,0)$ dynamically represent the participant’s running model-output baseline.

\textbf{Asynchronous Turn-Based Architecture:} To maintain natural interaction, the system must follow strict temporal constraints ($<2.0$s latency)~\cite{shiwa2009}. This conflicts with the computational cost of Generative Replay. We solve this using an \textit{Asynchronous Turn-Based Architecture} (Algorithm~\ref{Alg:solver_cl}) that separates interaction from learning.

\begin{itemize}

\item \textbf{Thread 1 (Interaction):} Operates on an inference time cycle. It performs inference using the current model snapshot ($\theta_{curr}$) to generate immediate counter-offers.

Because inference uses only a forward pass through the frozen encoder and a nearest-node match in the GWR, this step completes in under a second.

\item \textbf{Thread 2 (Learning):} Runs in the background. It dequeues interaction frames, generates ``Imagined’’ samples with the pretrained CAAE, and updates the model weights ($\theta_{train}$), sharing them with Thread 1 only at the end of each training step. Thread 2 never blocks Thread 1; if training exceeds the interaction cycle, the agent continues with the most recently completed GWR weights.

\end{itemize}

This asynchronicity enables the time constraint ($<2.0$s) to apply exclusively to the inference path (Thread~1). It completes a forward pass through the frozen encoder and nearest-node GWR match in under 200 ms per frame batch. Thread~2’s Generative Replay step, including CAAE sample generation and GWR weight changes, requires approximately 1–3 seconds per batch but runs asynchronously and never blocks inference.
The learning thread processes queued interaction data asynchronously. The GWR memories regulate adaptation using activation and habituation criteria. The regulated semantic-memory update additionally requires an $L_2$ discrepancy above $0.3$ between the incoming VA pseudo-label and the BMU's stored VA estimate.

\begin{algorithm}[ht]

\caption{Async. Turn-Based CL-enabled Solver Agent}
\label{Alg:solver_cl}
\begin{algorithmic}[1]
\Require Video Stream $S$, Pre-trained Weights $\theta_{init}$, CAAE $G$
\Ensure Bid Strategy $TU_{next}$

\State $\textbf{Init: } \theta_{curr}, \theta_{train} \leftarrow \theta_{init}; \quad \mathcal{Q}_{batch} \leftarrow \emptyset \quad \mathstrut $
\State $\textbf{Stats: } \boldsymbol{\mu}\leftarrow\mathbf{0},\ \boldsymbol{\sigma}\leftarrow\mathbf{1},\ \tilde E_0\leftarrow\mathbf{0}$

\Statex \textbf{Thread 1: Interaction Loop (Real-Time)}
\While{Negotiation Active}
\If{$\theta_{shared} \neq \text{null}$}
\State $\theta_{curr} \leftarrow \theta_{shared}; \quad \theta_{shared} \leftarrow \text{null}$
\EndIf
\State $\mathcal{B}_{t} \leftarrow \text{CaptureFrames}(S)$
\State $E_t\leftarrow\text{Infer}(\mathcal B_t;\theta_{curr})$
\State $\tilde E_t\leftarrow\text{Normalize}(E_t,\boldsymbol{\mu},\boldsymbol{\sigma})$
\State $\text{UpdateRunningStats}(\boldsymbol{\mu},\boldsymbol{\sigma},E_t)$
\State $\Delta_V \leftarrow \tilde{V}_t - \tilde{V}_{t-1}; \quad \Delta_A \leftarrow \tilde{A}_t - \tilde{A}_{t-1}$

\State $P_{sign} \leftarrow +1$ if $\Delta_V+\Delta_A\geq0$, else $-1$
\State $\delta_{det}\leftarrow\det(\tilde E_{t-1},\tilde E_t);\ \delta_{dot}\leftarrow\tilde E_{t-1}\cdot\tilde E_t$
\State $P_{rot} \leftarrow \operatorname{atan2}(\delta_{det}, \delta_{dot})$
\State $P_{rad} \leftarrow \big|\|\tilde{E}_t\|_2 - \|\tilde{E}_{t-1}\|_2\big|$
\State $(P_{rot},P_{rad})\leftarrow\text{ScaleFeedback}(P_{rot},P_{rad})$

\State $P_E \leftarrow P_{sign} \times \min\{1,\sqrt{P_{rot}^2+P_{rad}^2}\}$
\State $TU_{next} \leftarrow \text{SolverStrategy}(P_E, P_A, \Delta U)$
\State $\mathcal{Q}_{batch}\text{.Enqueue}(\mathcal{B}_{t})$
\EndWhile

\Statex \textbf{Thread 2: Continual Learning (Background)}
\While{Negotiation Active}
\If{$\mathcal{Q}_{batch} \neq \emptyset$}
\State  $\mathcal{D}_{t} \leftarrow \mathcal{Q}_{batch}\text{.Dequeue}()$
\State $\mathcal D_{gen}\leftarrow\text{GenerateReplay}(G)$
\State $\mathcal{D}_{train} \leftarrow \text{Mix}(\mathcal{D}_{t}, \mathcal{D}_{gen})$
\State $\theta_{train} \leftarrow \text{TrainStep}(\theta_{train}, \mathcal{D}_{train})$
\State $\theta_{shared} \leftarrow \theta_{train}$
\EndIf
\EndWhile
\end{algorithmic}
\end{algorithm}
\section{Experimental Setup}

\label{sec:experimentalSetup}

We investigate whether the personalized facial expression analysis module, through continual learning, significantly affects negotiation outcomes.
We also evaluate participants’ offer behavior and their interaction with the agent.

\subsection{Participants}

\label{subsec-participants}

We recruited 66 university students (39 male, 27 female; median age: 21) from \"{O}zye\u{g}in University, T\"{u}rkiye. All participants were educated in English, with academic levels ranging from undergraduate (BS) to doctoral (PhD) studies. The human-agent experiments in this study were approved by the Ethics Committee of \"{O}zye\u{g}in University, and written informed consent was obtained from all participants.
\subsection{Embodied Agent Architecture}

\begin{figure}[ht]
\centering
\begin{tabular}{cc}
\subfloat[Participant view]{\label{fig-nao-front}\includegraphics[width=0.46\columnwidth]{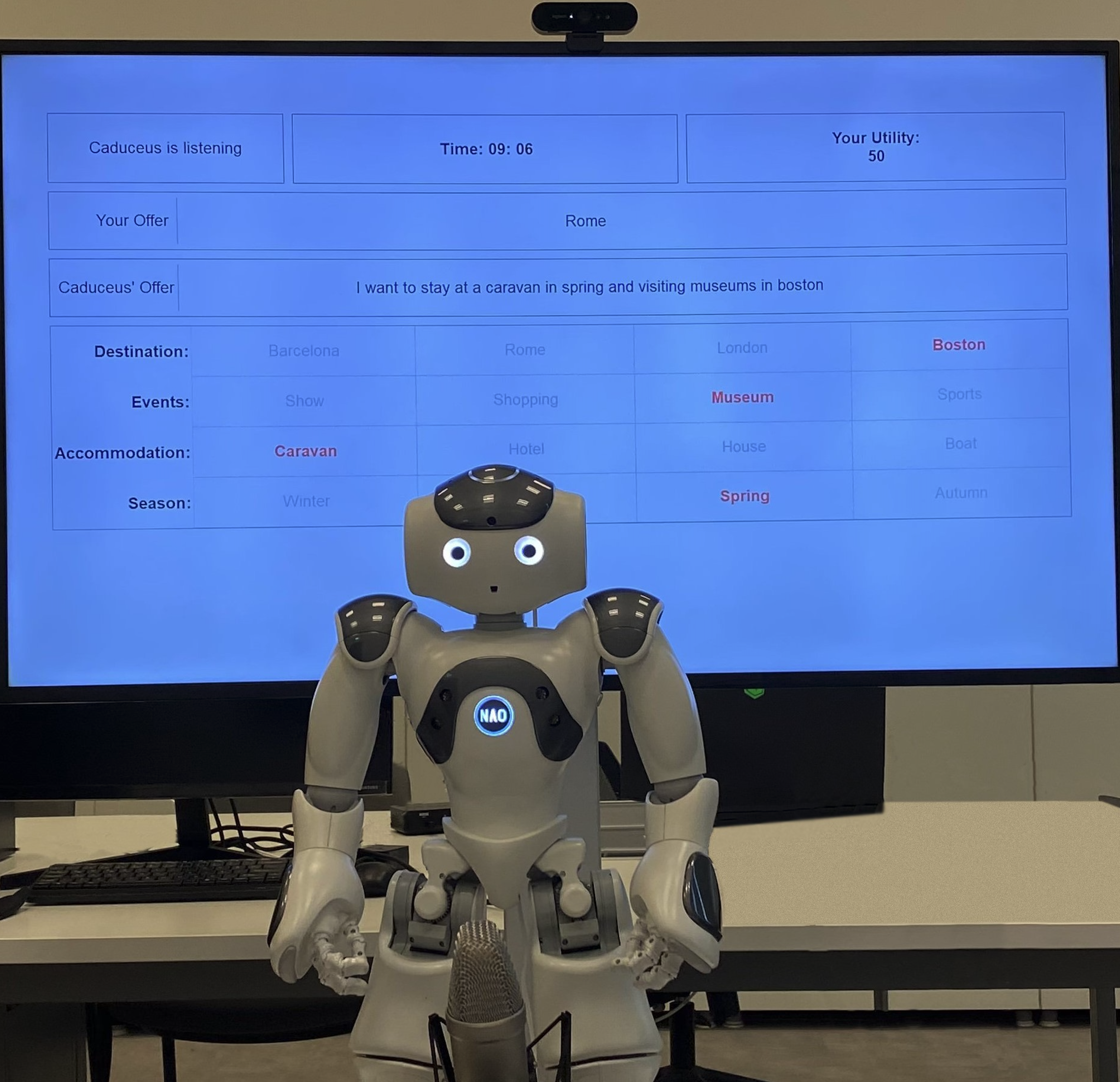}}&
\subfloat[Side view]{\label{fig-virtual-exp-setup}\includegraphics[width=0.445\columnwidth]{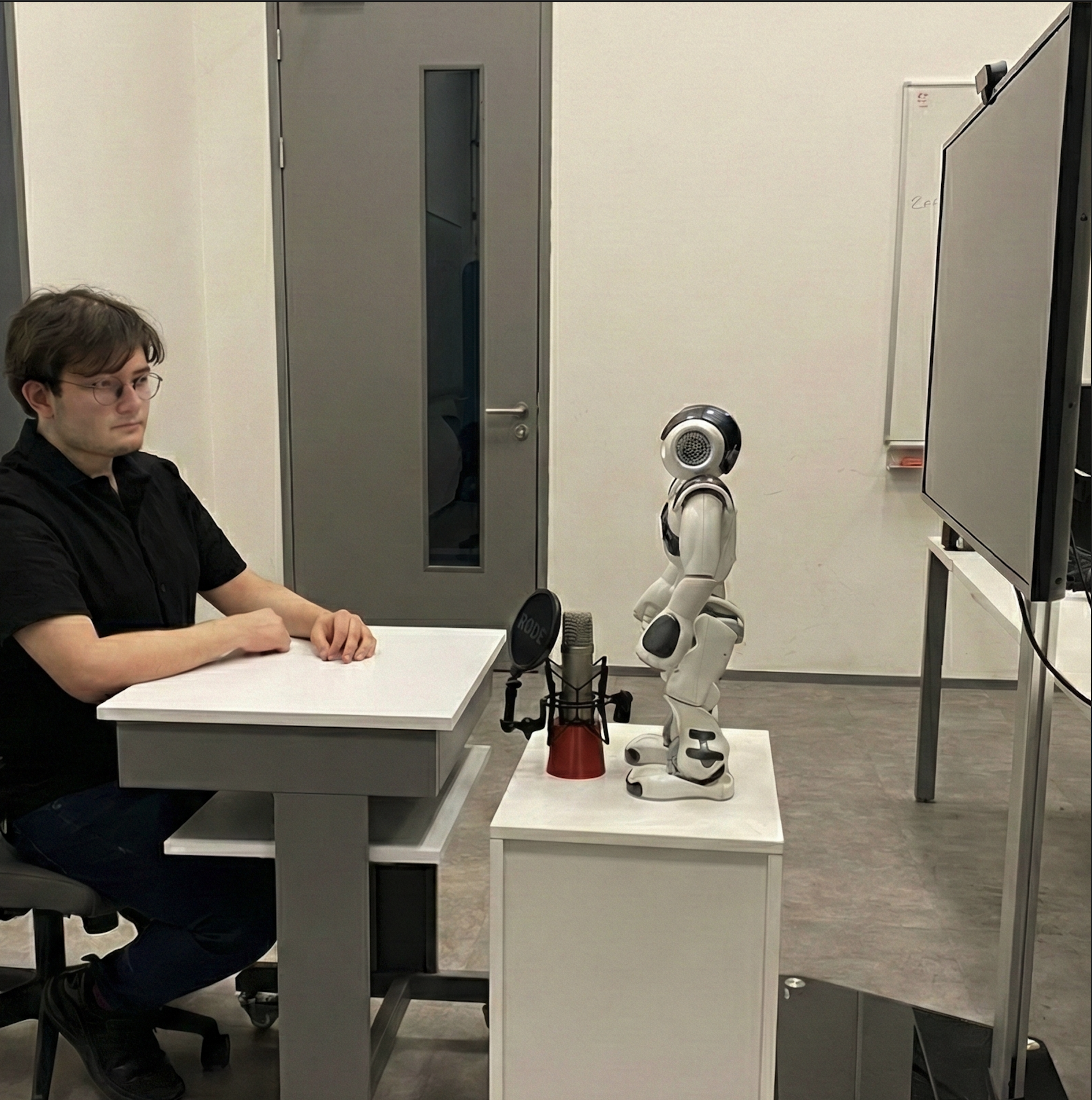}}\
\end{tabular}
\caption{Human-Robot Negotiation Experimental Setup: (a) Participant's view of the Nao robot and negotiation interface. (b) Side view of the room.}
\label{fig-naosetup}
\end{figure}

To validate our approach with an embodied agent in a complex negotiation domain, we utilize the Negotiator framework \cite{negotiator2024}. This modular architecture supports embodied negotiation. As shown in \cref{fig:architecture}, the Negotiator orchestrates components that handle the alternating offer protocol, analyze preferences, and manage multimodal interaction modules.

These include visual, verbal, and gesture generation, all via humanoid robots such as Nao, Pepper, and QT, or via virtual avatars.
The Nao robot embodied the Solver agent, as shown in \cref{fig-naosetup}.
As shown in Fig.~\ref{fig-nao-front}, participants view the Solver interface on a large screen.
This interface displays the ordered alternative values for each issue according to participants' preferences, the utility of the current offer, the current chat offers for both sides, and the time elapsed during negotiation.

The participant starts by verbally specifying an offer with a value for each issue. The Speech-to-Offer Module processes the offer, and the free speech protocol also lets the participant make arguments. The framework's modularity allows us to swap the standard perception component for our CL module. The Nao robot provided a physically present negotiation partner, drawing on the role of embodiment in human-robot interaction \cite{lee2006}. We ensured a consistent appearance to avoid the confounding effects noted by Keskin et al. \cite{keskin2024}. Each side knows only its own preferences during the negotiation. If the parties cannot reach an agreement, both sides receive zero utility.

We captured facial expressions with a Logitech Brio 4K webcam at 60 fps, positioned at eye level and 1.2-1.5\,m from the participant. We capped per-round frame counts at 90 for CLIFER processing. The system ran on a laptop with an NVIDIA GTX 1070, 32 GB of RAM, and a 1 TB SSD.

\subsection{Experiment Procedure}

\begin{figure}[ht]
\centerline{\includegraphics[width=\columnwidth]{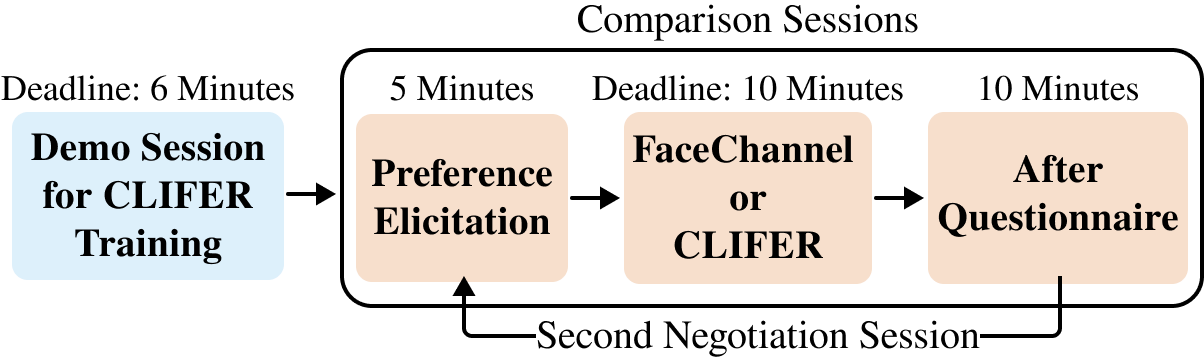}}
\caption{Experiment Procedure: A 6-min Grocery demo initializes CLIFER. Participants then complete two Holiday negotiations (10-min deadline), one with each perception model in counterbalanced order, each preceded by preference elicitation and followed by a questionnaire.}
\label{fig:exp_procedure}
\end{figure}

As shown in \cref{fig:exp_procedure}, each participant negotiated under two facial-expression perception conditions: the generic FaceChannel model and the continual-learning CLIFER model. 
After a demo session in a separate domain (Section~\ref{subsec:demo}), participants negotiated twice with the Solver Agent, following their self-ordered preferences.
We counterbalanced the order of the two perception conditions across participants. Both main negotiations used the Holiday domain with a ten-minute deadline (Fig.~\ref{fig:exp_procedure}). Section~\ref{sec:evaluation} examines order sensitivity.
After completing the first negotiation session, they took a 5-minute break before starting the second session. Immediately after each negotiation, participants completed a post-survey.

\subsubsection{Negotiation Domain}

Before each session, participants ranked the issues by importance and indicated their preferred values for that domain.
For the Holiday Domain shown in \cref{tab:holiday_values}, the participant earns a vacation gift from a draw. Another person they have never met also wins the same vacation. They must go on vacation together. They negotiate where to go (\textit{Destination}), where to stay (\textit{Accommodation}), when to go (\textit{Season}), and what to do (\textit{Events}). A participant can list Destination first, Accommodation second, and so on. They also order their preferences, such as London $\prec$ Boston $\prec$ Rome $\prec$ Barcelona, to show that they prefer Barcelona over Rome.

\begin{figure}[ht]
\centerline{\includegraphics[width=0.85\columnwidth]{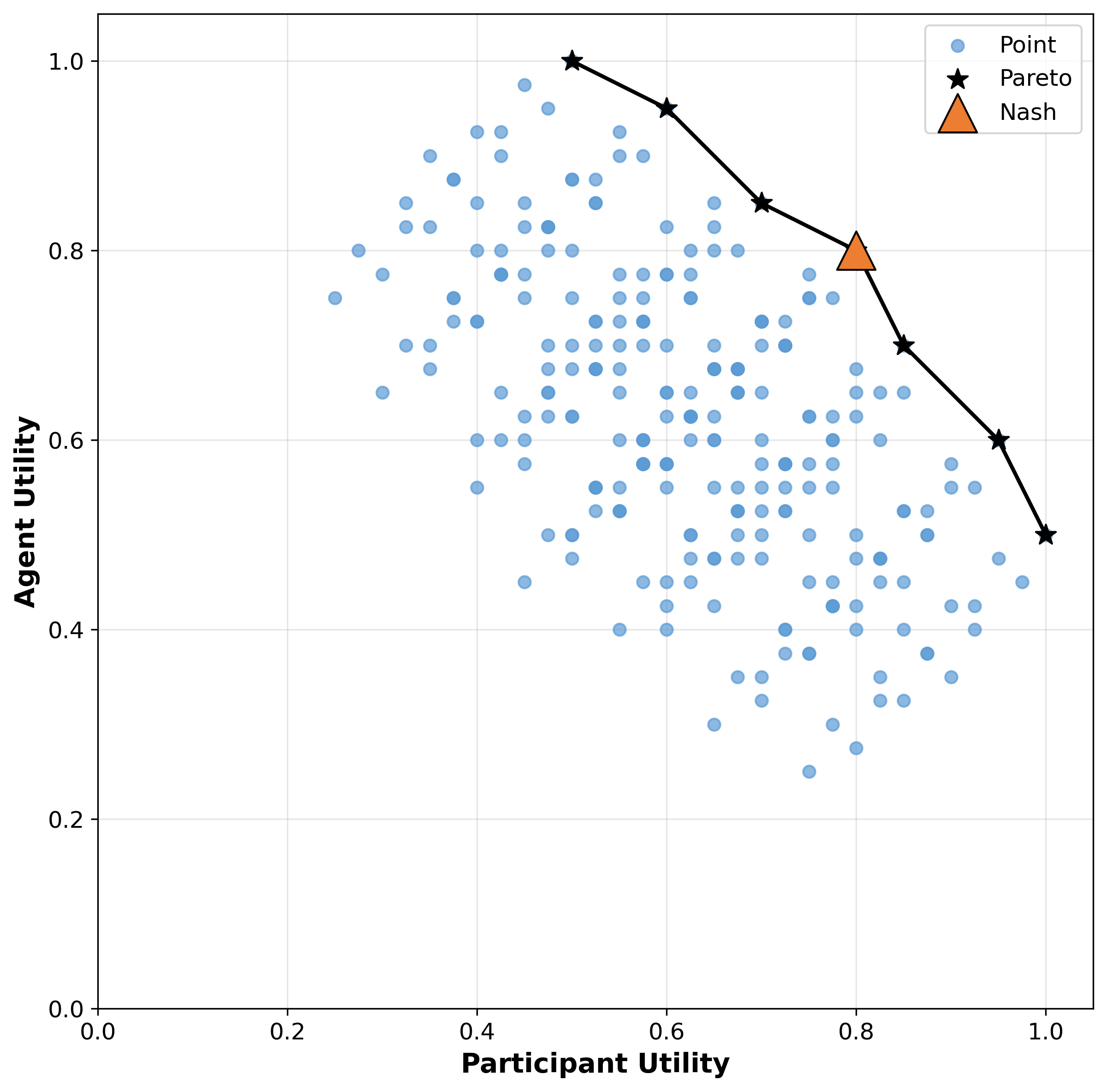}}
\caption{Utility Space of the Holiday Domain: Each point represents a possible agreement; stars mark the Pareto frontier and the triangle the Nash solution (0.80, 0.80). Both sessions use the same space.}
\label{fig:bid_space_exp}
\end{figure}

Using these preference rankings, the Negotiator core arranged the agent’s preferences to distribute the offers with the same utility distribution for each session, as shown in \cref{fig:bid_space_exp}. Both parties have equal negotiation power, and to reduce familiarity with specific offer alternatives, we differentiated the destination places for their second negotiation.

\begin{table}[ht]
\centering
\caption{Holiday Domain Preferences: \\ The Issue Weights \& Value Orders}
\label{tab:holiday_values}
\resizebox{\columnwidth}{!}{\begin{tabular}{lcccc}
\toprule
\textbf{Issue (Weight)} & \textbf{1} & \textbf{2} & \textbf{3} & \textbf{4} \\
\midrule
\textbf{Destin.} (0.48) & \makecell{Barcelona (1)\\Venice (2)} & \makecell{Rome (1)\\Lisbon (2)} & \makecell{London (1)\\Sydney (2)} & \makecell{Boston (1)\\Miami (2)} \\
\addlinespace
\textbf{Accom.} (0.32) & AirBnB & Hotel & Hostel & Boat \\
\addlinespace
\textbf{Events} (0.16) & Shopping & Museum & Sports & Show \\
\addlinespace
\textbf{Season} (0.04) & Summer & Winter & Spring & Fall \\
\bottomrule
\end{tabular}}
\end{table}

\subsubsection{Demo Session}
\label{subsec:demo}

A fundamental issue in applying CL to short-term interactions is the cold-start problem, where the model lacks enough initial data to define a participant’s baseline~\cite{Churamani2020}.
Before the main negotiations, each participant completed a six-minute practice negotiation with the robot in a separate \textit{Grocery} domain. This demo served two purposes. First, participants exchanged offers with the robot to become familiar with the interface and procedure without rehearsing the Holiday alternatives. Second, their facial samples, pseudo-labeled by the frozen FaceChannel model, initialized CLIFER's participant-specific episodic memory. We excluded demo observations from the main-session outcome analyses.

At the start of that participant's CL session, CLIFER loaded the model trained during the demo and continued adapting to newly captured facial samples. FaceChannel remained fixed in the FC session. We excluded FC-session facial expressions from CL training, including when FC came first. The comparison therefore evaluates demo-session training together with continued online adaptation (Section~\ref{sec:conclusion}).
\section{Evaluation}

\label{sec:evaluation}

We evaluate affective outputs, negotiation behavior, outcomes, and participant ratings. The utility and concession comparisons assess the strategy's practical value; the signal analyses describe its affective input. We collected data from 66 participants, each completing two main human-agent negotiations (132 sessions).

The principal comparisons used paired participant summaries. We used two-tailed paired $t$-tests and reported paired Cohen's $d_z$ (the mean CL-FC difference divided by its sample standard deviation) to estimate effect size. Following Cohen's conventions, we interpret the absolute effect size as negligible below $0.2$, small from $0.2$ to below $0.5$, medium from $0.5$ to below $0.8$, and large at $0.8$ or above. We used an initial significance threshold of $\alpha=0.05$. To control for multiple testing across our four principal comparisons (estimated arousal, concession rate, agent utility, and agreement rounds), we applied the Holm-Bonferroni sequential correction procedure. We also report a separate post hoc Holm correction across the six participant move categories. Questionnaire, affect-by-move, correlation, and mediation analyses remain exploratory and fall outside the four-outcome correction. Tables report unadjusted $p$-values; adjusted values are given in the text.

Post hoc checks of outcome uncertainty used paired $t$ intervals, 10,000 percentile bootstrap resamples, and leave-one-participant-out estimates. Resampling retained both sessions per participant and, for round-level checks, all their rows. The reported intervals are unadjusted 95\% CIs. Participant-level values are means and sample SDs; paired effect sizes use $d_z$. Robot-move comparisons used paired $t$-tests with Holm correction across six categories. For presentation-order sensitivity, we regressed within-person CL-FC differences on presentation order using HC3 standard errors. Holm correction covered five outcomes (arousal, concessions, both utilities, and rounds). First-session contrasts used Welch tests.

We counterbalanced the experiment order (CL-first: $N=34$, FC-first: $N=32$), distributing session exposure across conditions. Equal weighting of the two orders retained positive agent-utility ($0.022$, 95\% CI $[0.004,0.040]$) and concession-rate differences ($0.091$, $[0.019,0.164]$). Order checks assess differences between sequences, separately from the principal condition contrasts; only agreement rounds showed order sensitivity after the five-outcome Holm correction ($p_{adj}=0.048$). First-session-only utility and concession estimates were smaller and less precise, with intervals including zero. Table~\ref{tab:sequence_outcomes} gives means and sample SDs by session and sequence. Counterbalancing handles common session effects; condition-specific carryover cannot be excluded.

For context, a sensitivity calculation using G*Power~\cite{faul2007gpower} indicates approximately 80\% power to detect $d_z=0.35$ with $N=66$ paired samples and $\alpha=0.05$. The secondary signal analyses used $2,114$ logged rows nested within $132$ sessions from $66$ participants. Treating these rows as independent would cause pseudo-replication and inflate Type 1 error rates.

Secondary analyses of round-level affect estimates used linear mixed-effects models (LMMs) with participant random intercepts, allowing participants to have different baselines. To check the pooled CL-FC contrasts, we added random session intercepts within participants, allowing each person's two sessions to have different baselines. A separate regression check used participant fixed effects and standard errors clustered by participant, allowing observations from the same person to be correlated when estimating uncertainty. We used the intraclass correlation coefficient (ICC) to summarize between-participant variation.

\begin{figure*}[!t]
\centering
  \begin{minipage}[t]{0.49\textwidth}
  \vspace{5pt}   \centering
      \begin{minipage}{\linewidth}
\centering
\subfloat[ \textbf{Subject-Level Affective Signal:} CL significantly increases mean estimated arousal ($d_z=0.68$) at the subject level. \label{fig:emo_boxplot_subject}]{\includegraphics[width=\linewidth]{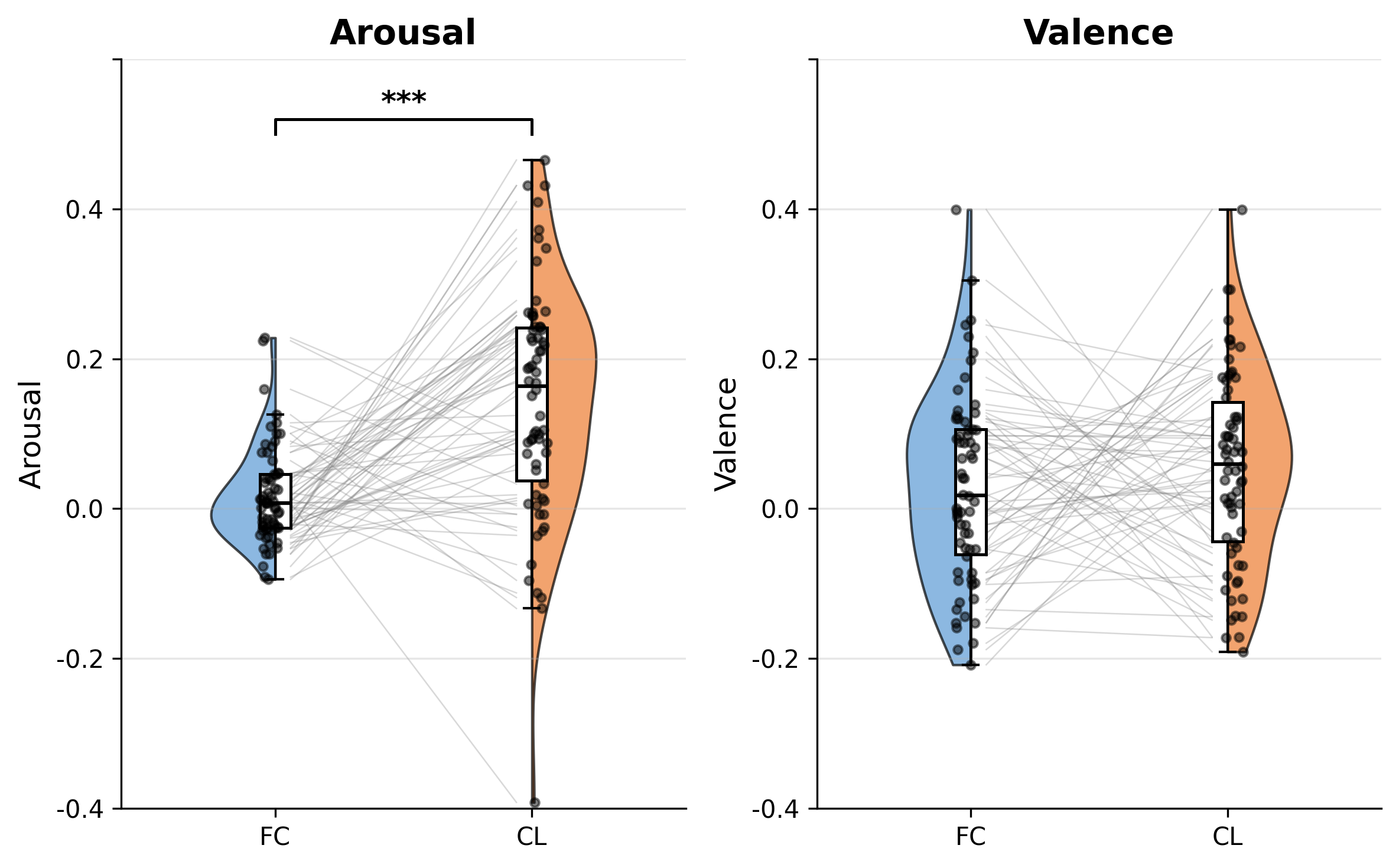}}

      \vspace{0.50cm}
      \end{minipage}
      \begin{minipage}{\linewidth}
\centering
\subfloat[ \textbf{Round-Level Affective Signal:} Pooled arousal ($d=0.88$) and valence contrasts; stars indicate nominal tests. Participant-block inference supports only the arousal contrast. \label{fig:emo_boxplot_round}]{\includegraphics[width=\linewidth]{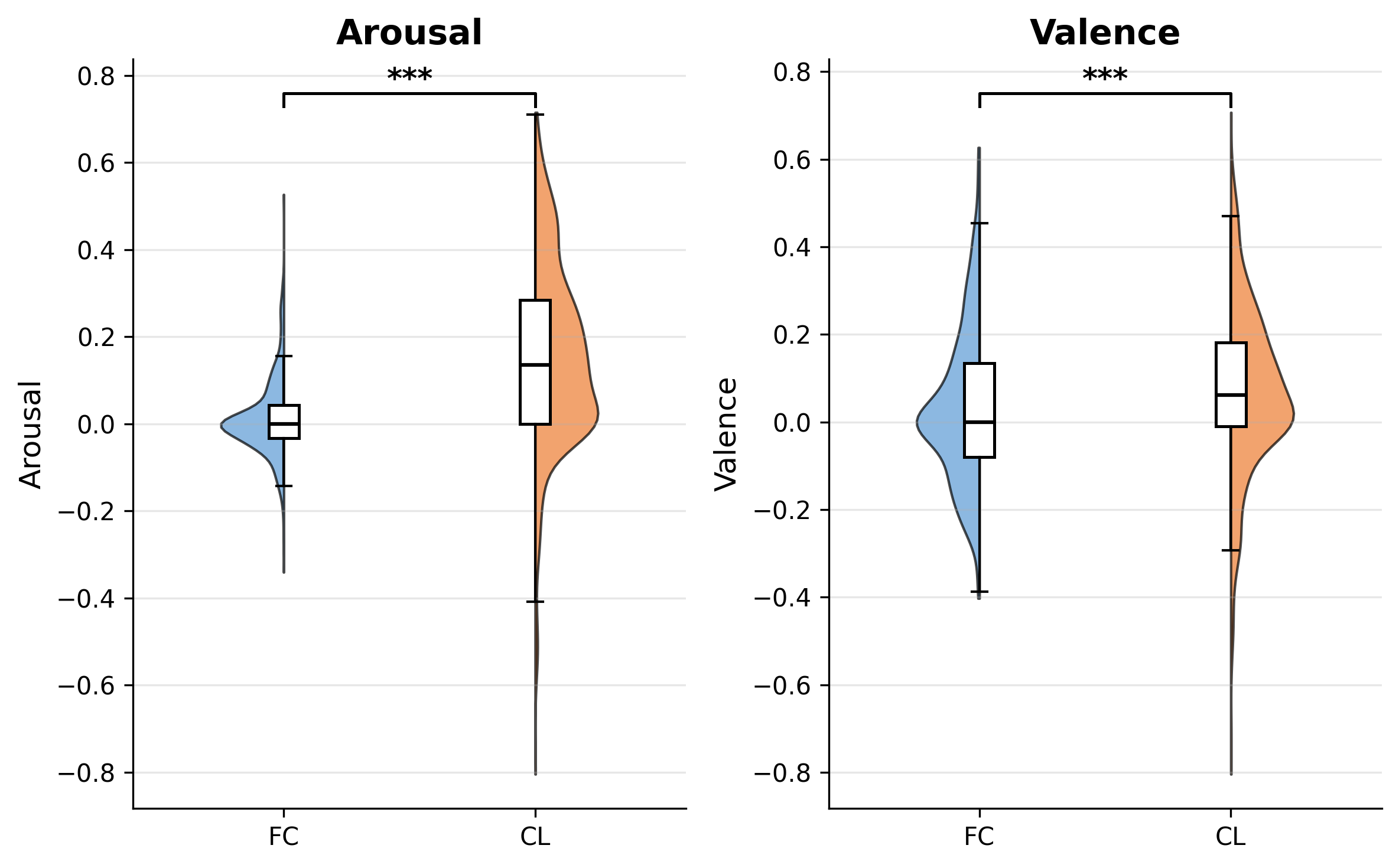}}

      \end{minipage}
      \vspace{0.5cm}
      \end{minipage}
  \hfill   \begin{minipage}[t]{0.49\textwidth}
  \vspace{10pt}   \centering
      \begin{minipage}{\linewidth}
\centering
\subfloat[ \textbf{Individual Affective Difference Analysis:} 75.8\% of subjects show higher arousal estimates with CL (** $p < 0.01$); 51.5\% show higher valence estimates. \label{fig:emo_responder}]{\includegraphics[width=\linewidth]{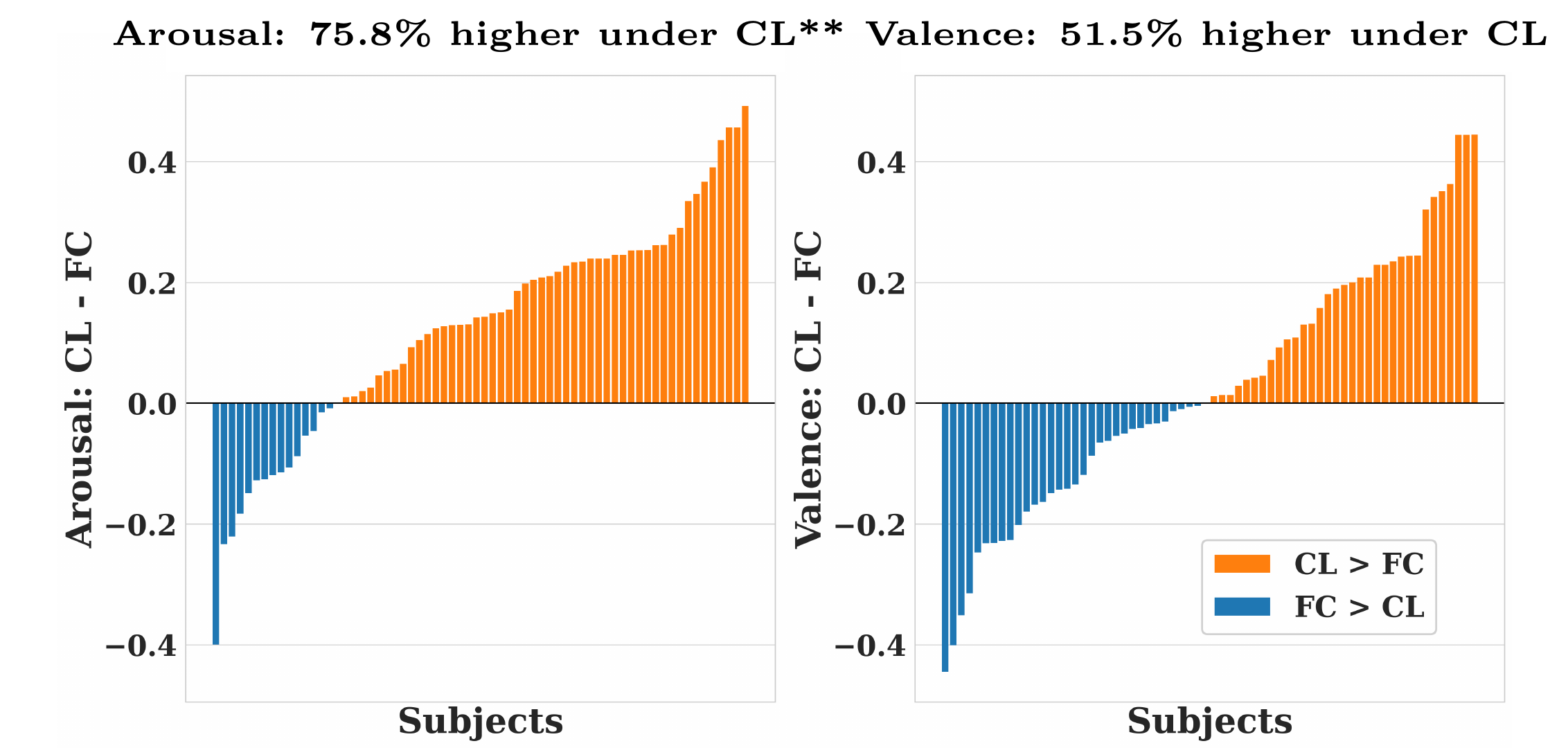}}

      \vspace{1.0cm}
      \end{minipage}
      \begin{minipage}{\linewidth}
\centering
\subfloat[ \textbf{Temporal Profiles in Sessions:} CL has a higher mean estimated arousal after the initial zero entries; the first 30 logged offer positions are shown. \label{fig:emo_temporal}]{%
\includegraphics[width=\linewidth]{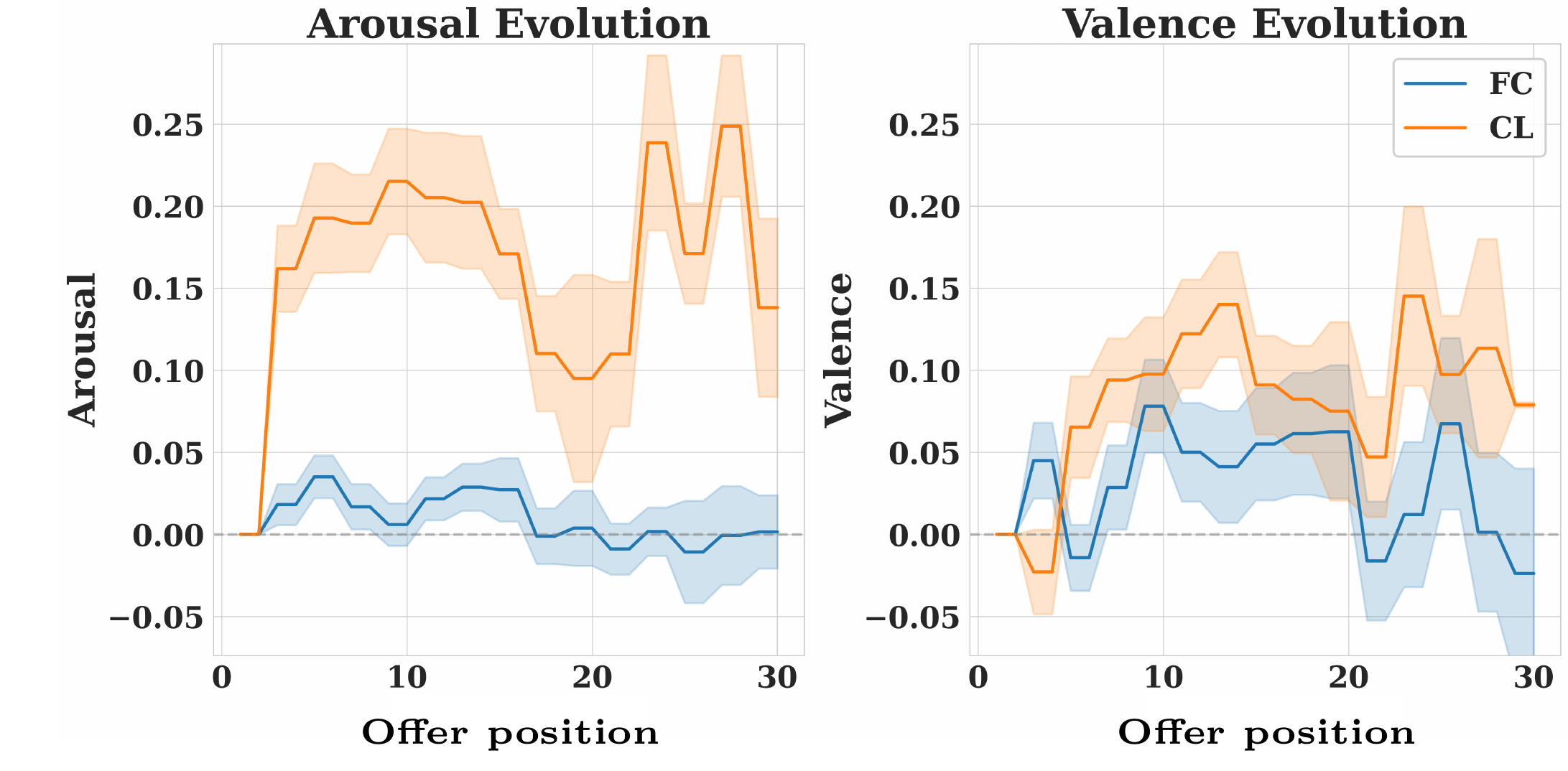}}

      \end{minipage}
      \end{minipage}
      
  \vspace{5pt}   \begin{minipage}{0.32\textwidth}
\centering
\captionsetup[subfloat]{justification=centering}
\subfloat[\textbf{Round-Level} ($n_{FC}=1{,}134$, $n_{CL}=980$) \\ ($r$: FC=$-0.15$, CL=$\mathbf{0.45}$)\label{fig:corr_round}]{\includegraphics[width=\linewidth]{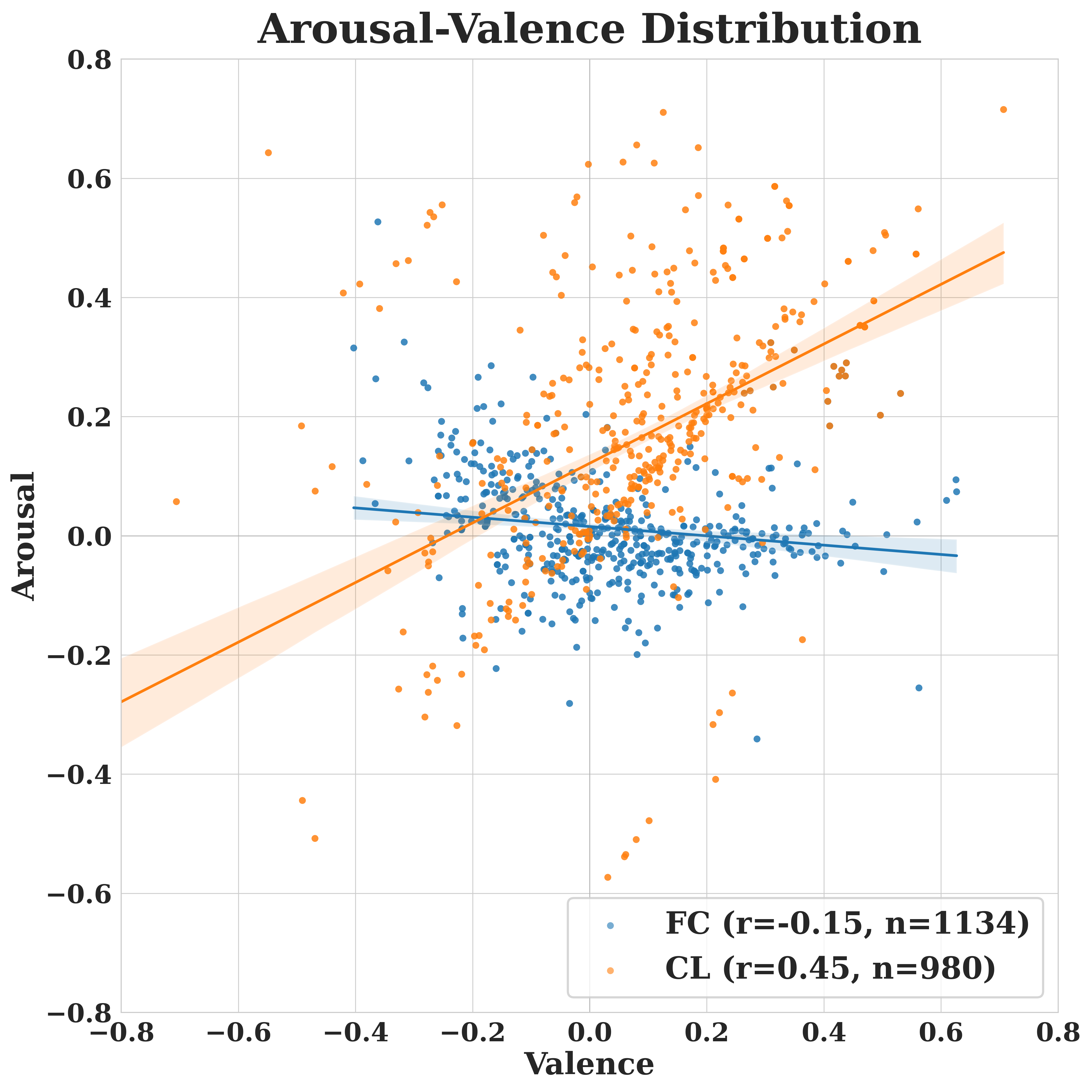}}

  \end{minipage}
  \hfill
  \begin{minipage}{0.32\textwidth}
\centering
\captionsetup[subfloat]{justification=centering}
\subfloat[\textbf{Balanced Sample} ($n=980$ per condition) \\ ($r$: FC=$-0.14$, CL=$\mathbf{0.45}$) \label{fig:corr_balanced}]{\includegraphics[width=\linewidth]{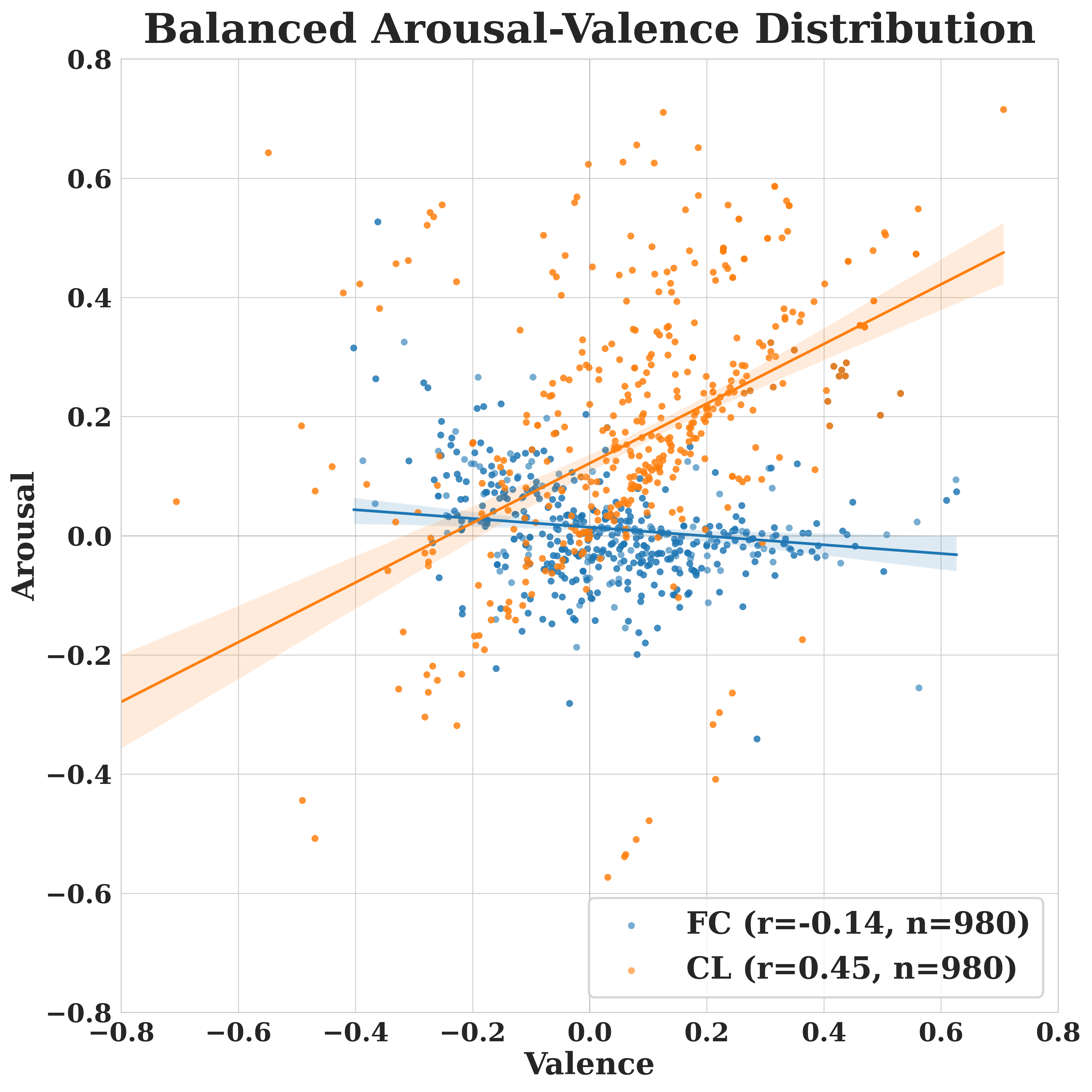}}

  \end{minipage}
  \hfill
  \begin{minipage}{0.32\textwidth}
\centering
\captionsetup[subfloat]{justification=centering}
\subfloat[\textbf{Subject-Level} ($N=66$) \\ ($r$: FC=$-0.20$, CL=$\mathbf{0.31}$)\label{fig:corr_subject}]{\includegraphics[width=\linewidth]{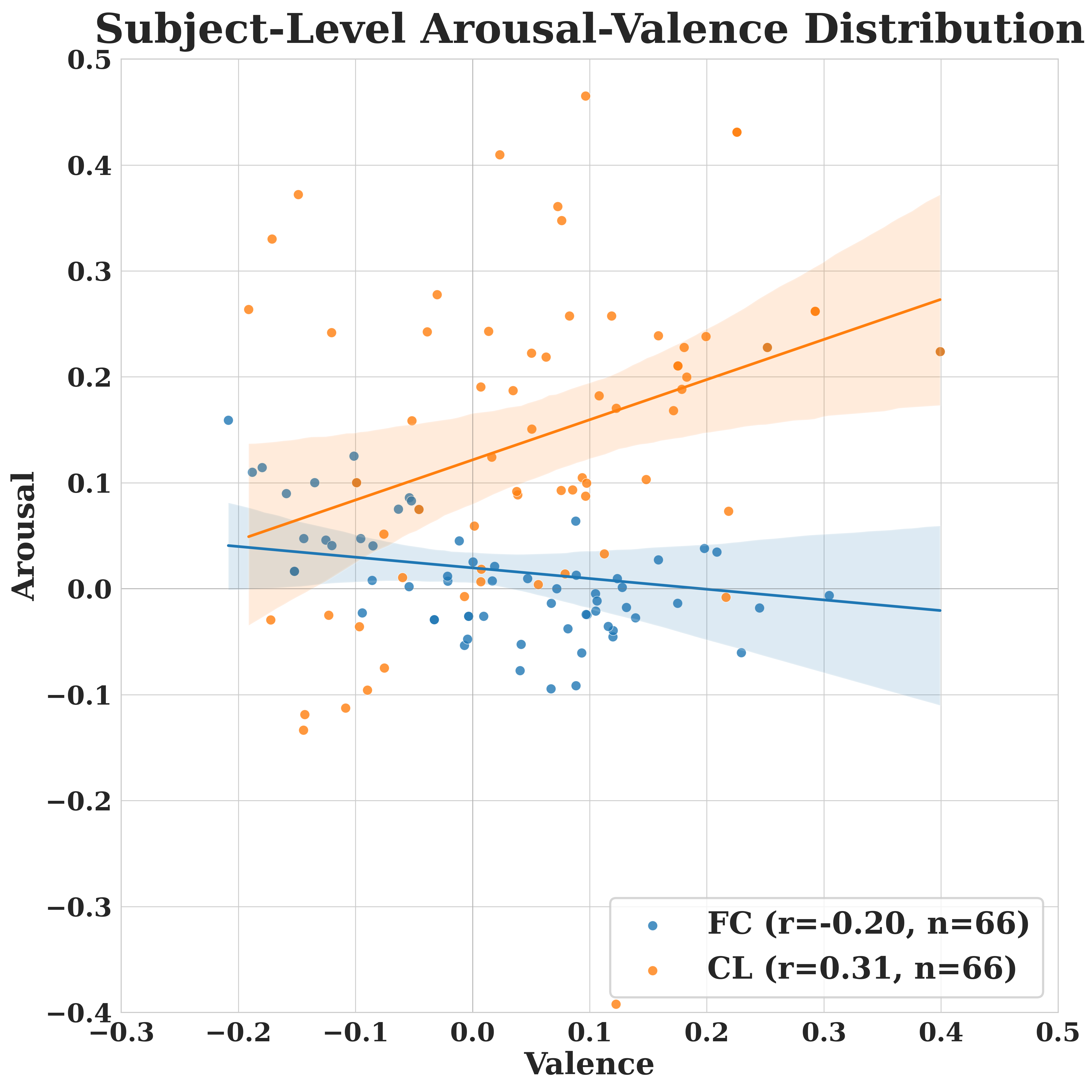}}
  \end{minipage}
\caption{ \textbf{Affect Perception Analysis:} (a-b) \textbf{Affective Signal:} CL produces significantly higher arousal estimates at both the subject ($d_z=0.68$) and round levels ($d=0.88$). (c) \textbf{Individual Differences:} Most subjects ($75.8\%$) showed higher arousal values with the CL-enabled Solver Agent. (d) \textbf{Temporal Profile:} The offer-indexed means include initial zero entries and fewer contributing sessions at later positions. (e-g) \textbf{Correlations:} CL produces a positive arousal-valence correlation ($r=0.45$ at round level; $r=0.31$ at subject level) compared with the negative association seen with FaceChannel at round and subject levels. }
\label{fig:emotion_structure_composite}
\end{figure*}

\subsection{Affective Feedback}

\label{sec:affective_perception}

To evaluate the effectiveness of the continual learning approach in affect perception (\textbf{RQ1}), we analyzed normalized arousal and valence predictions over all 66 participants.
Each participant completed sessions with both the generic FaceChannel (FC) and adaptive CLIFER (CL) models, yielding $2,114$ rows in the exported round-level dataset. We used two analysis levels: Subject-Level and Round-Level. Participant-level paired tests provide the primary comparisons ($N=66$); secondary LMMs and round-level plots describe the signal structure.

\subsubsection{Affective Signal Coherence \& Sensitivity}

Personalizing the agent's perception expanded the range of its estimated arousal signal (Fig.~\ref{fig:emo_boxplot_subject}). At the subject level, CL produced significantly higher mean arousal estimates than FC, yielding a medium effect size (Cohen's $d_z=0.68$, $p<0.001$). Table~\ref{tab:emotion_comprehensive} reports the means and dispersions at each aggregation level. The secondary round-level analysis ($n=2{,}114$) showed the same direction ($d=0.88$). An equal-size subsample comparison ($n=980$ per condition) also retained this contrast ($d=0.87$); balancing row counts addresses sample-size differences, but does not remove within-session dependence.

The arousal contrast remained when the LMM accounted for both participant and session baselines. The model estimated higher arousal under CL than FC (CL-FC: $\beta=0.127$, 95\% CI $[0.087,0.168]$). For valence, the estimated difference was $\beta=0.027$ (95\% CI $[-0.017,0.072]$), with an interval including zero. Regressions with participant fixed effects and standard errors clustered by participant supported the same conclusions for both dimensions. Bootstrap resampling of whole participants, retaining both sessions and all their rows, also supported the pooled arousal difference ($\Delta=0.142$, 95\% CI $[0.095,0.186]$). A sensitivity check excluding the two initial zero-valued entries per session retained all 66 participant pairs; the arousal difference remained positive ($\Delta=0.146$, 95\% CI $[0.092,0.199]$; paired $t(65)=5.45$, $p<0.001$, $d_z=0.67$).

The CL outputs also had greater variance (ratio $=5.7$; pooled Levene's $F=518.97$, nominal $p<0.001$). This pooled test does not account for dependence among repeated observations. These results describe the range of model outputs used by the negotiation controller. This comparison does not separate perceptual adaptation from condition-specific output scaling.

While the pooled valence comparison was nominally significant ($d=0.18$, $p<0.001$), it did not reach significance at the subject level ($p=0.317$). The bootstrap interval for the pooled valence difference included zero ($\Delta=0.033$, $[-0.021,0.087]$), so its nominal round-level significance was not robust to this dependence check. Valence also remained non-significant after excluding the initial zero entries ($p=0.383$). Thus, arousal showed the clearer contrast.

Collapsing consecutive identical VA estimates per session resulted in 925 logged rows; descriptive correlations remained positive for CL ($r=0.44$) and negative for FC ($r=-0.16$). This checks log repetitions, not unique video frames.

\subsubsection{Individual Affective Difference Analysis}

The overall increase in estimated arousal reflects differences across many participants rather than a few extreme outliers (Fig.~\ref{fig:emo_responder}). Tracking each participant's exact CL$-$FC difference ($\Delta$) revealed that CL yielded higher arousal estimates by an average of $+0.125$ normalized units (95\% CI [0.080, 0.170]). We classified participants by the magnitude of this output difference into four descriptive categories: \textbf{Strong} ($\Delta > 0.15$) comprising 47\% of subjects; \textbf{Moderate} ($0.05 < \Delta \leq 0.15$) capturing 21\% of subjects; \textbf{Neutral} ($|\Delta| \leq 0.05$) representing 14\% of subjects; and \textbf{Negative} ($\Delta < -0.05$) at 18\%. The bars in Fig.~\ref{fig:emo_responder} show signed differences; these categories apply the thresholds above.

Overall, 75.8\% of participants showed higher arousal estimates with the CL model, a proportion significantly above 50\% (binomial test, $p<0.01$). Valence differences varied across participants; the paired mean comparison was not significant ($p=0.317$, Table~\ref{tab:emotion_comprehensive}). The \textbf{Negative} group (12/66 participants, 18\%) had already exhibited unusually high arousal scores under the baseline FC model. This variation suggests that the effect of adaptation on the signal depends on the participant and interaction context.

\subsubsection{Temporal Change}

The CL-FC arousal difference varies across the negotiation timeline (Fig.~\ref{fig:emo_temporal}). The horizontal index follows logged offers, including two initial zero-valued entries in each session; successive offers can share an affect estimate. After these entries, CL has higher mean estimated arousal across the displayed positions. Later means are based on fewer sessions, because only negotiations still in progress contribute at each offer position.

\subsubsection{Arousal-Valence Coupling}

Personalization changes the feedback coupling between arousal and valence. Fig.~\ref{fig:emotion_structure_composite}e shows that the CL condition exhibits a moderate positive correlation ($r=0.45$, $n=980$), so that higher estimated valence tends to accompany higher estimated arousal.
In contrast, the generic FC model shows a weak negative correlation ($r=-0.15$, $n=1{,}134$).
The two conditions therefore provide coupled feedback signals in different ways.

\begin{table}[ht]
\centering
\caption{Estimated Affect Statistics}
\label{tab:emotion_comprehensive}
\small
\begin{tabular*}{\columnwidth}{@{\extracolsep{\fill}}lcccc@{\extracolsep{0pt}}}
   \toprule

   \textbf{Metric} & \textbf{FC} & \textbf{CL} & \textbf{\textit{p}} & \textbf{\textit{d}} \\
   \midrule
   \multicolumn{5}{l}{\textit{Subject-Level ($N=66$; $d_z$)}} \\

Arousal & $0.02 \pm 0.07$ & $\mathbf{0.14 \pm 0.16}$ & $\mathbf{<0.001}$ & \textbf{0.68} \\

Valence & $0.03 \pm 0.13$ & $0.05 \pm 0.13$ & 0.317 & 0.12 \\
   \midrule
   \multicolumn{5}{l}{ \textit{Round-Level ($n_{FC}=1{,}134$, $n_{CL}=980$; $d$)}} \\
   Arousal & $0.01 \pm 0.09$ & $\mathbf{0.15 \pm 0.22}$ & $\mathbf{<0.001}$ & \textbf{0.88} \\
   Valence & $0.03 \pm 0.18$ & $0.07 \pm 0.20$ & $<0.001$ & 0.18 \\
   \midrule
   \multicolumn{5}{l}{ \textit{Balanced Sample ($n=980$ per condition; $d$)}} \\
   Arousal & $0.01 \pm 0.09$ & $\mathbf{0.15 \pm 0.22}$ & $\mathbf{<0.001}$ & \textbf{0.87} \\
   Valence & $0.03 \pm 0.17$ & $0.07 \pm 0.20$ & $<0.001$ & 0.18 \\
   \bottomrule
\end{tabular*}
\end{table}

FC sessions often required more rounds to reach agreement, resulting in slightly imbalanced observation counts ($n_{FC} > n_{CL}$). To examine the influence of unequal observation counts, we tested an equal-size subsample (Fig.~\ref{fig:emotion_structure_composite}f), where both groups had the same number of observations ($n=980$ per condition). Notably, the correlation in CL remained positive ($r=0.45$), whereas in FC it remained negative ($r=-0.14$). The contrast persists after equalizing row counts.

This coupling difference persists at the subject level (Fig.~\ref{fig:emotion_structure_composite}g).
The correlations among the 66 participant means are positive for CL ($r=0.31$) and negative for FC ($r=-0.20$). We averaged within-session arousal-valence correlations in Fisher-$z$ space and back-transformed the means: $r=0.88$ for CL and $r=-0.27$ for FC.
A paired-sample comparison of Fisher's $z$-transformed per-subject correlations confirms the CL–FC coupling difference is highly significant ($t(65)=7.73$, $p<0.001$, $d_z=0.95$) and uses participants as the units of comparison. A positive association is compatible with engagement in mutually favorable exchanges, but the VA framework does not require positive coupling as a criterion for cooperation or accurate perception~\cite{russell1980}. These results address the signal-level component of \textbf{RQ1}; the human-agent negotiation outcomes below assess its functional value.

\begin{figure*}[!t]
\centering
\begin{minipage}[b]{0.48\textwidth}
\centering
\subfloat[FC \label{fig-circumplex-move-fc}]{\includegraphics[width=\textwidth]{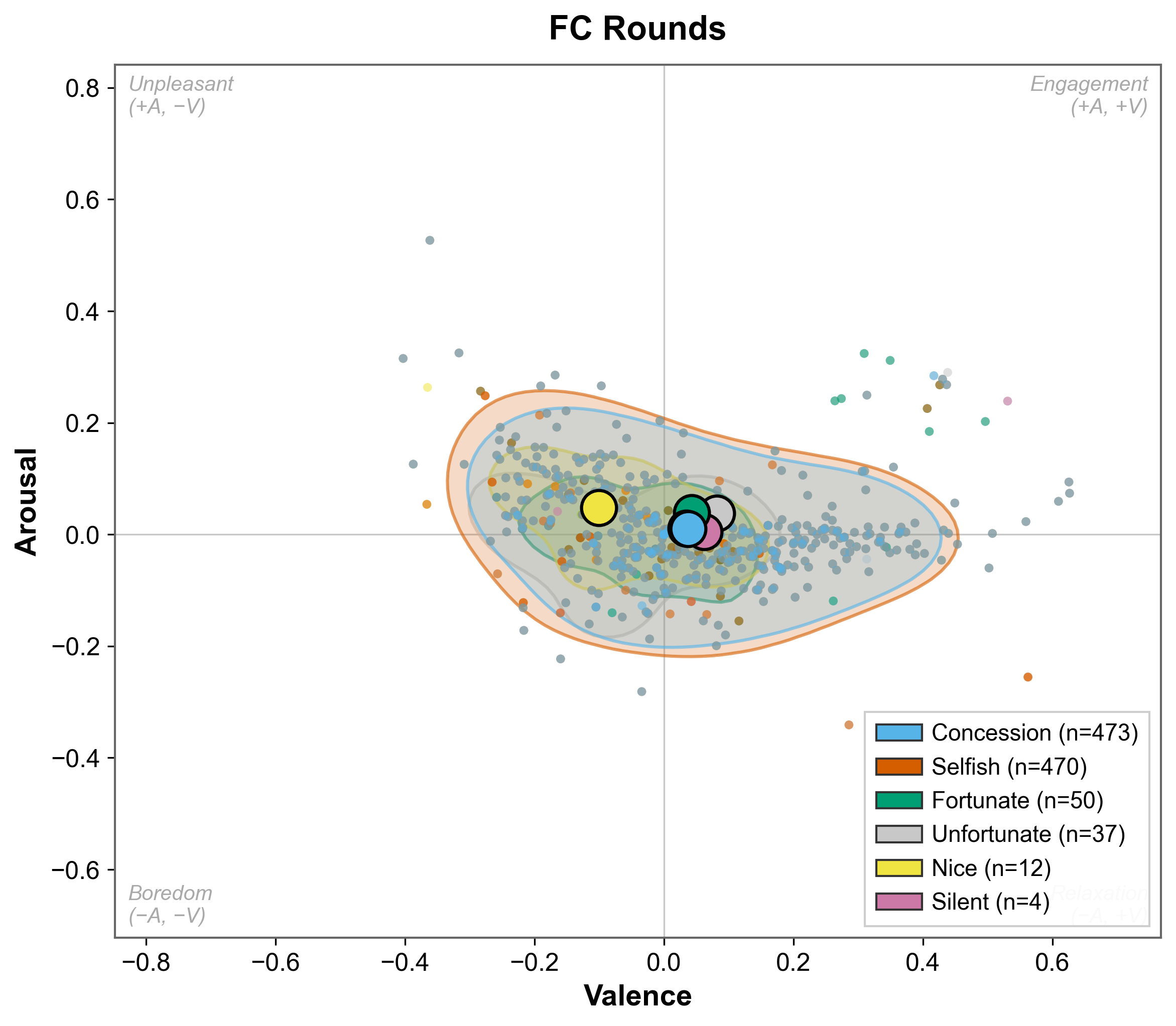}}

\end{minipage}
\hfill
\begin{minipage}[b]{0.48\textwidth}
\centering
\subfloat[CL \label{fig-circumplex-move-cl}]{\includegraphics[width=\textwidth]{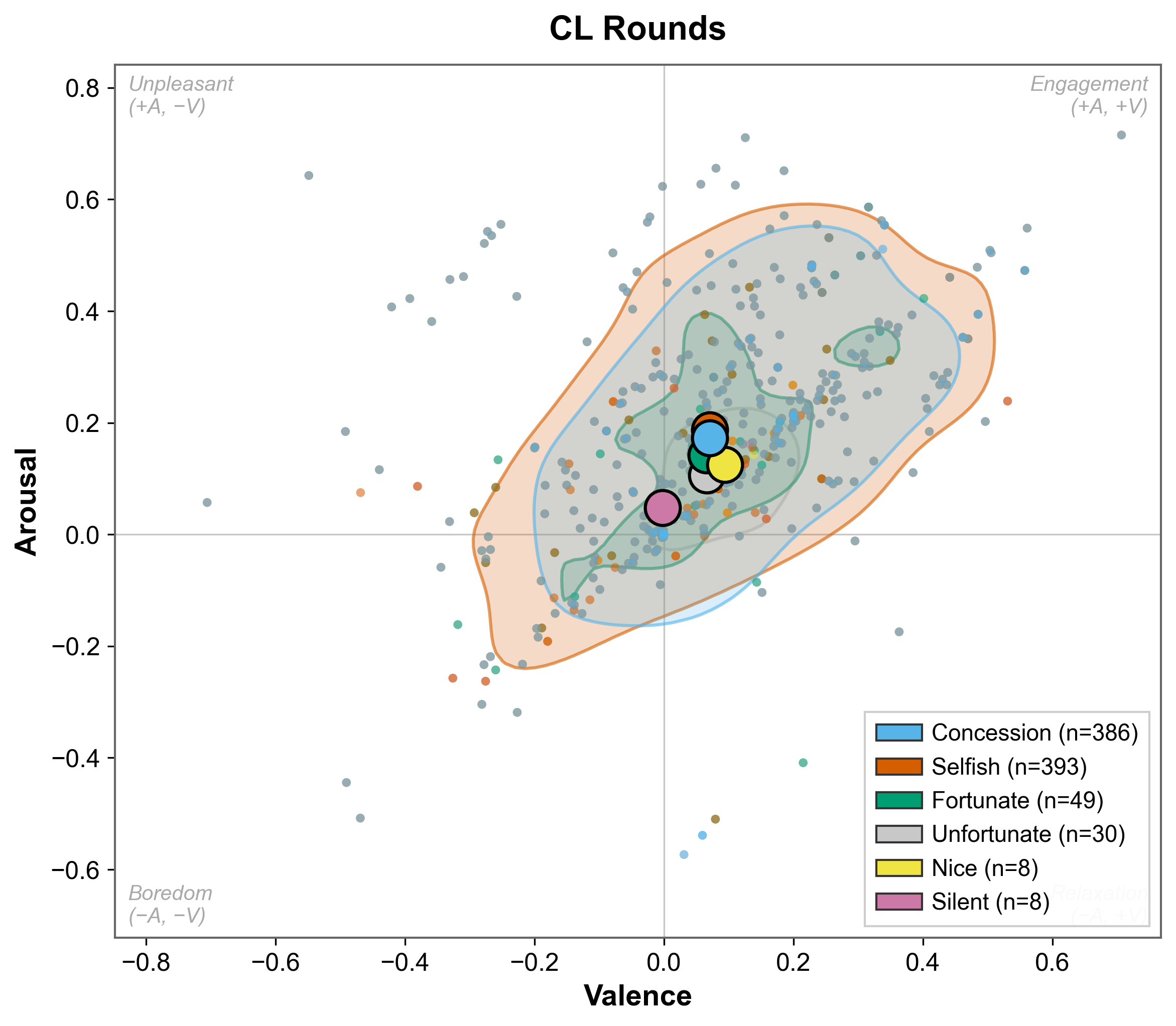}}

\end{minipage}
\caption{Estimated Affect by Move Type: Round-level VA estimates for (a) FC and (b) CL, pooling human and robot moves; colored regions show density and large circles denote per-move centroids. CL centroids are higher in arousal across all move types.}
\label{fig-circumplex-move}
\end{figure*}

\subsection{Behavioral Influence}
\label{sec:behavioral_influence}

The CL-enabled Solver agent uses personalized affective feedback to regulate concessions in response to the opponent's offers and time pressure. This lets the strategy respond to changes in the participant-adapted signal throughout the interaction.

\begin{figure}[tb]
\centering
\includegraphics[width=\columnwidth]{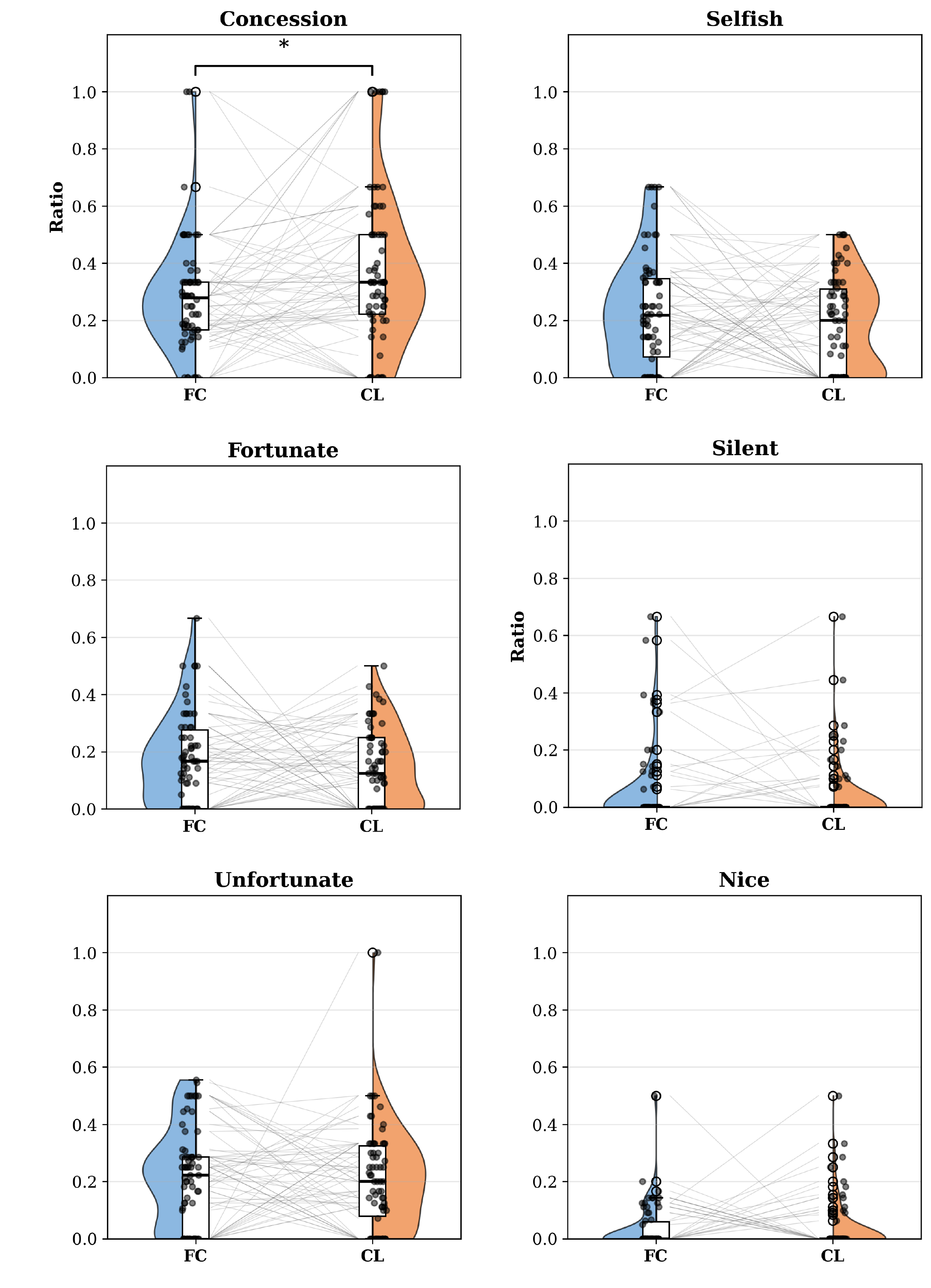}
\caption{Participant Offer Moves: Per-participant proportions under FC and CL; lines connect each participant's two sessions. Participants made a higher proportion of concession moves under CL (unadjusted * $p<0.05$).}
\label{fig:move_distributions}
\end{figure}

\begin{table}[ht]
\centering
\caption{Participant Offer Moves}
\label{tab:moves_stats}
\small
\begin{tabular*}{\columnwidth}{@{\extracolsep{\fill}}lcccc@{\extracolsep{0pt}}}
\toprule

\textbf{Metric} & \textbf{FC} & \textbf{CL} & \textbf{\textit{p}} & \textbf{$d_z$} \\
\midrule

\textbf{Concession} & $0.28 \pm 0.20$ & $\mathbf{0.37 \pm 0.27}$ & \textbf{0.013} & \textbf{0.31} \\

Nice & $0.04 \pm 0.08$ & $0.04 \pm 0.09$ & 0.821 & 0.03 \\

Fortunate & $0.18 \pm 0.16$ & $0.15 \pm 0.14$ & 0.272 & -0.14 \\

Unfortunate & $0.21 \pm 0.17$ & $0.21 \pm 0.18$ & 0.803 & -0.03 \\

Selfish & $0.23 \pm 0.19$ & $0.19 \pm 0.17$ & 0.172 & -0.17 \\
Silent & $0.06 \pm 0.14$ & $0.04 \pm 0.11$ & 0.390 & -0.11 \\
\bottomrule
\end{tabular*}
\end{table}

Participants showed a significant behavioral shift toward concession (Fig.~\ref{fig:move_distributions}, Table~\ref{tab:moves_stats}). Specifically, participants who interacted with the CL agent increased their concession rate from 28.1\% to 37.4\% ($p=0.013$, $p_{adj}=0.040$, $d_z=0.31$). This result is compatible with the idea that participants interpret the agent's reactions when adjusting their offers, as suggested by the EASI framework~\cite{vanKleef2009}. It may also reflect persuasion or adaptation to the agent's bargaining stance; concession rates alone do not distinguish these explanations from cooperation. Meanwhile, the rate of \textit{Unfortunate} moves, which reduce both parties' utilities, did not change significantly between conditions. A separate post hoc Holm correction across all six participant move categories gave $p_{adj}=0.0795$ for concession; the move-category comparisons remain exploratory.

Applying the same move taxonomy to successive offers by the same actor reproduced the participant concession summaries. Rates exclude the first offer, which has no initial move. Robot concession rates were 38.9\% in both conditions (63 pairs; CL-FC 95\% CI $[-7.81,7.82]$ percentage points; $p_{adj}=1.000$). Three CL sessions contained only one robot offer, so we could not define a transition rate. This check did not detect lower aggregate robot concession frequency under CL.

\subsubsection*{Mapping Affect to Negotiation Behavior}

This exploratory affect-by-move analysis pools logged human and robot moves. Its nominal round-level tests do not fully represent temporal dependence or identify actor-specific affective reactions. To explore the behavioral shift, we mapped the agent's estimated affective feedback onto logged negotiation moves (Fig.~\ref{fig-circumplex-move}, Table~\ref{tab:move_arousal}). We distinguish two comparisons: (1) differences in the estimated signal between CL and FC, and (2) differences associated with move types within each condition (e.g., Concession vs. Selfish). The move label describes the offer-maker, whereas the affect estimate always comes from the participant's face.

To assess whether the condition difference extends across move types, we compared estimated arousal between CL and FC within each reported category. Notably, the CL model produced higher arousal estimates than the static FC baseline for all major move types. The largest arousal contrasts occurred during the most frequent actions: Concession ($d=1.05$) and Selfish ($d=1.10$); together, these account for most negotiation moves. Furthermore, the other listed move types also showed higher CL estimates (Fortunate $d=0.58$, Nice $d=0.89$, Unfortunate $d=0.34$). As shown in Fig.~\ref{fig-circumplex-move}b, the circumplex space shows clear upward centroid shifts. Overall, these exploratory comparisons indicate that the signal difference extends across several move categories.

\begin{table}[ht]
\centering
\caption{Estimated Arousal by Move Type}
\label{tab:move_arousal}
\small
\begin{tabular*}{\columnwidth}{@{\extracolsep{\fill}}lcccc@{\extracolsep{0pt}}}
\toprule
\textbf{Move Type} & \textbf{FC Arousal} & \textbf{CL Arousal} & \textbf{\textit{p}} & \textbf{\textit{d}} \\
\midrule
\textbf{Concession} & $0.011$ & $\mathbf{0.173}$ & $\mathbf{<0.001}$ & \textbf{1.05} \\
\textbf{Selfish} & $0.011$ & $\mathbf{0.187}$ & $\mathbf{<0.001}$ & \textbf{1.10} \\
\textbf{Nice} & $0.048$ & $\mathbf{0.125}$ & \textbf{0.001} & \textbf{0.89} \\
\textbf{Fortunate} & $0.039$ & $\mathbf{0.142}$ & $\mathbf{<0.001}$ & \textbf{0.58} \\
Unfortunate & $0.038$ & $0.107$ & $0.018$ & 0.34 \\
\bottomrule
\end{tabular*}
\end{table}

We fitted linear mixed-effects models (LMMs) to examine how estimated arousal varies by logged move type within each condition. In the CL condition, moves that increased the counterpart’s utility (Fortunate: $\beta=-0.054$, $p<0.001$; Concession: $\beta=-0.051$, $p=0.004$) had lower estimated arousal than Selfish moves (the Selfish baseline). These coefficients describe within-condition associations between logged move type and estimated arousal. In the corresponding FC analysis, the nominal comparison indicated a difference only for Unfortunate moves ($\beta=-0.023$, $p=0.002$). Separate condition-specific fits do not test whether the move-related associations differ between conditions. The substantial between-subject variance observed here (ICC $\approx$ 0.53-0.55) indicates between-participant variation under the fitted models. Together, the circumplex mapping and LMMs connect \textbf{RQ1} and \textbf{RQ2} by showing how the estimated affective signal relates to negotiation moves. They complement the concession-rate finding.

\subsection{Negotiation Efficiency}

\begin{figure}[ht]
\includegraphics[width=\columnwidth]{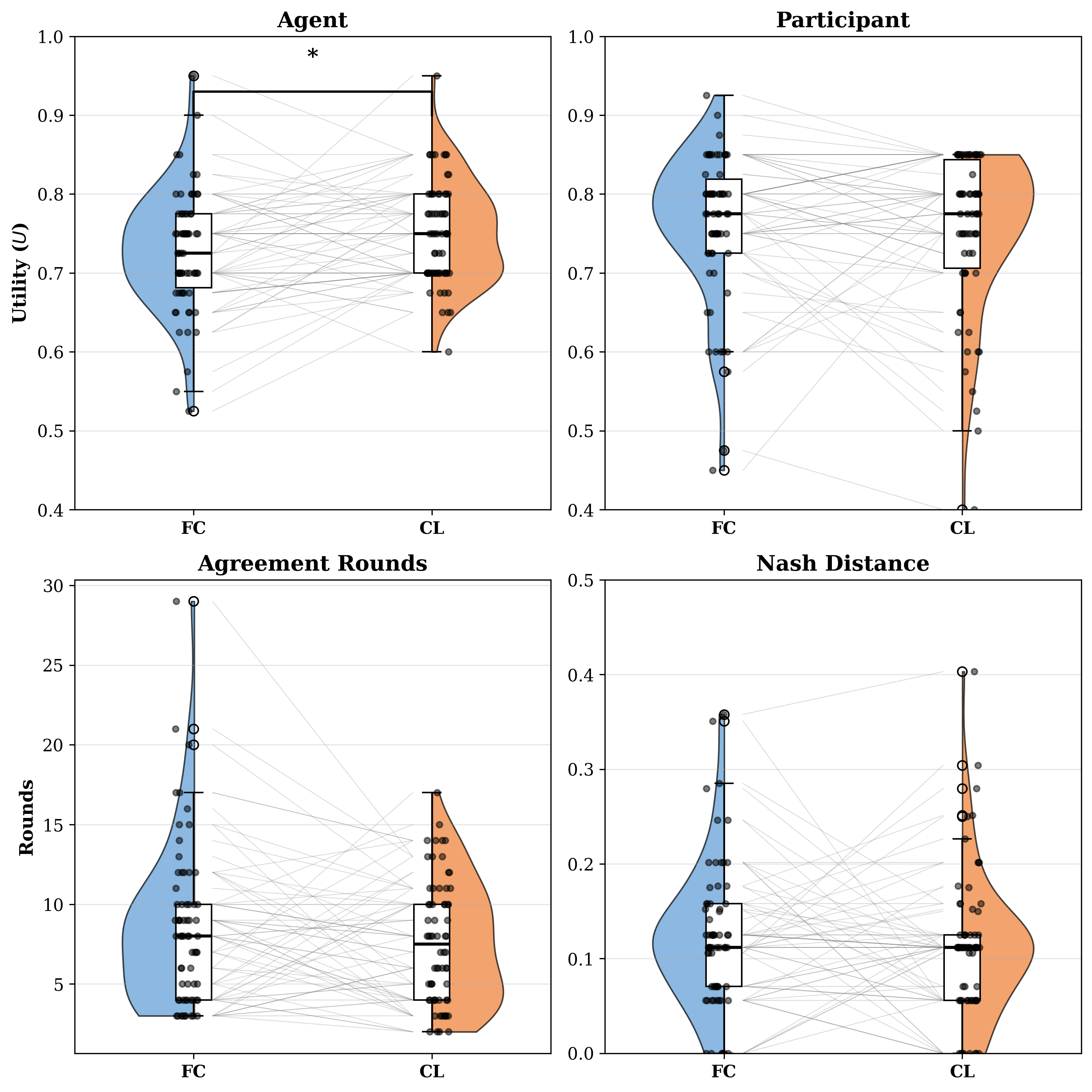}
\centering
\caption{Negotiation Outcomes: Paired raincloud plots compare agent utility, participant utility, agreement rounds, and Nash distance. CL yielded significantly higher agent utility (unadjusted * $p<0.05$); the other differences were not statistically significant.}
\label{fig-scores}
\end{figure}

As illustrated in Fig.~\ref{fig-scores} and Table~\ref{tab:utility_stats}, the CL-enabled Solver Agent achieved a significantly higher final utility than the generic FC baseline (\textbf{RQ3}; raw $p=0.017$, $p_{adj}=0.040$, $d_z=0.30$). Participant Utility ($p=0.495$) and overall Nash distance ($p=0.378$) did not show significant differences between conditions. Furthermore, the CL-enabled agent reached agreements in 11\% fewer rounds on average than the FC baseline, although this reduction was not statistically significant ($p=0.134$). Together, the outcomes show higher agent utility, with no statistically significant difference in participant utility or proximity to the reference Nash solution. The paired CL-FC differences were $+0.022$ in agent utility (95\% CI $[0.004,0.040]$), $-0.008$ in participant utility ($[-0.031,0.015]$), and $+9.35$ percentage points in participant concessions ($[2.02,16.69]$). Participant-level bootstrap intervals also excluded zero for the agent-utility and concession gains; both directions persisted across all 66 leave-one-participant-out checks.

\begin{table}[ht]
\centering
\caption{Negotiation Outcomes}
\label{tab:utility_stats}
\small
\begin{tabular*}{\columnwidth}{@{\extracolsep{\fill}}lcccc@{\extracolsep{0pt}}}
\toprule

\textbf{Metric} & \textbf{FC} & \textbf{CL} & \textbf{\textit{p}} & \textbf{$d_z$} \\
\midrule

\textbf{Agent Utility} & $0.73 \pm 0.08$ & $\mathbf{0.75 \pm 0.06}$ & \textbf{0.017} & \textbf{0.30} \\

Participant Utility & $0.76 \pm 0.09$ & $0.75 \pm 0.10$ & 0.495 & -0.08 \\
Nash Distance & $0.12 \pm 0.08$ & $0.11 \pm 0.08$ & 0.378 & -0.11 \\

Agree. Rounds & $8.56 \pm 5.05$ & $7.61 \pm 3.85$ & 0.134 & -0.19 \\
\bottomrule
\end{tabular*}
\end{table}

\begin{table}[ht]
\centering
\caption{Outcomes by Session and Sequence}
\label{tab:sequence_outcomes}
\small
\setlength{\tabcolsep}{1pt}
\begin{tabular*}{\columnwidth}{@{\extracolsep{\fill}}lcccc@{\extracolsep{0pt}}}
\toprule
Metric & {\scriptsize Session 1: CL} & {\scriptsize Session 2: FC} & {\scriptsize Session 1: FC} & {\scriptsize Session 2: CL} \\
\midrule
Est. Arousal & $0.17{\pm}0.16$ & $0.00{\pm}0.05$ & $0.03{\pm}0.07$ & $0.11{\pm}0.15$ \\
Concession Rate & $38.8{\pm}27.1$ & $22.1{\pm}18.7$ & $34.5{\pm}18.9$ & $36.0{\pm}27.8$ \\
Agent Utility & $0.75{\pm}0.07$ & $0.72{\pm}0.09$ & $0.74{\pm}0.06$ & $0.75{\pm}0.06$ \\
Rounds & $7.56{\pm}3.85$ & $10.09{\pm}5.47$ & $6.94{\pm}4.06$ & $7.66{\pm}3.92$ \\
\bottomrule
\end{tabular*}
\end{table}

\subsection{Questionnaire Analysis}

\begin{figure}[ht]
\centering
\includegraphics[width=\columnwidth]{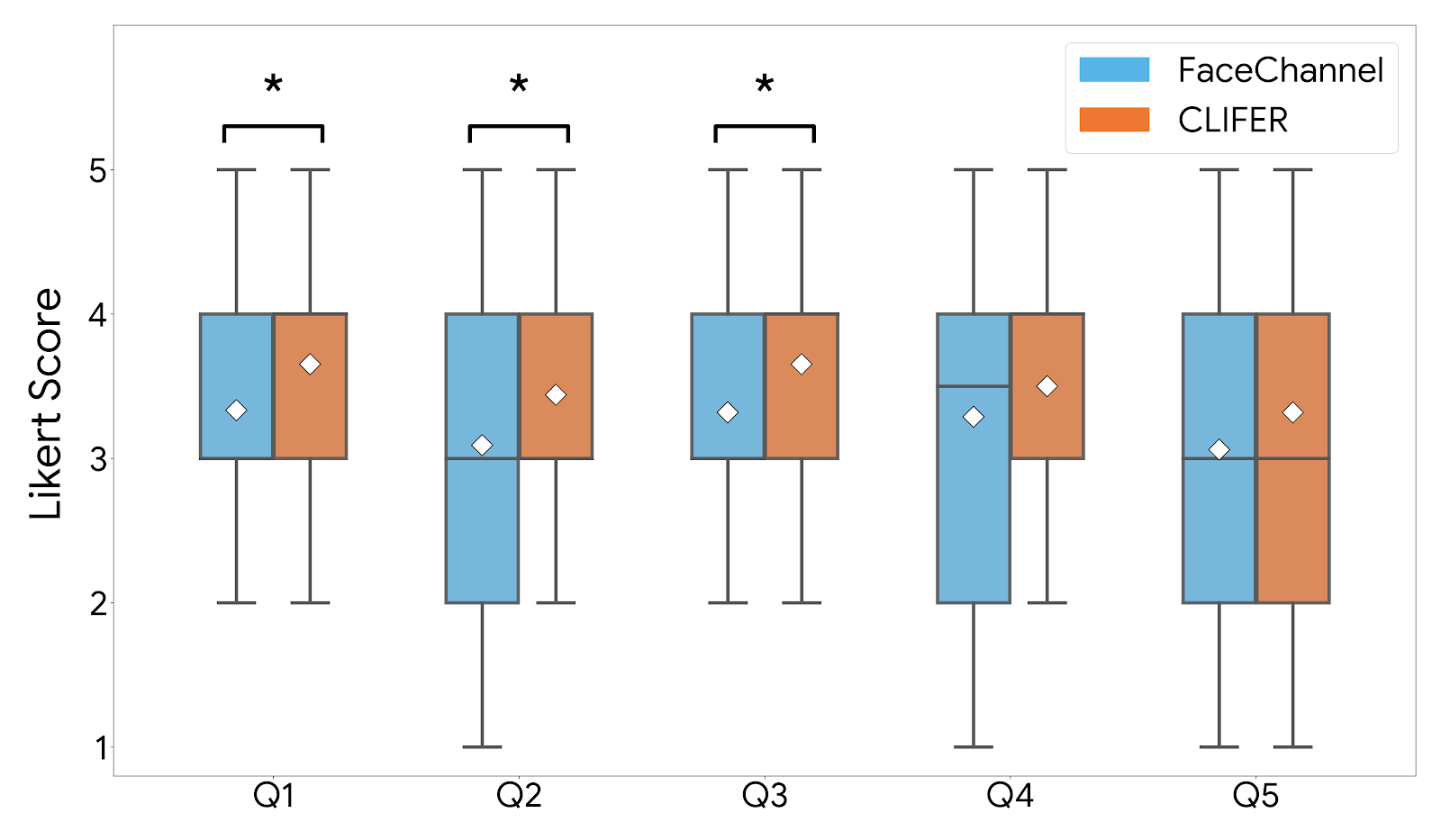}
\caption{Questionnaire Responses: Ratings on a 5-point Likert scale for Q1–Q5 (Table~\ref{tab:questionnaire_stats}); white diamonds mark means. CL received higher ratings on Q1–Q3 (unadjusted * $p<0.05$, exploratory).}
\label{fig:comparative-questions}
\end{figure}

Personalizing the agent’s affective perception also changed its subjective evaluation (\textbf{RQ4}). As shown in Fig.~\ref{fig:comparative-questions} and detailed in Table~\ref{tab:questionnaire_stats}, participants rated the CL-enabled agent higher on preference modeling, affective awareness, and persuasiveness. These exploratory comparisons concern single questionnaire items, rather than validated multi-item constructs. Both conditions shared the negotiation policy, although their affective inputs could produce different offers.

\begin{table}[ht]
\centering

\caption{Questionnaire Responses: Two-tailed Wilcoxon signed-rank tests; paired Cohen's $d_z$.}
\label{tab:questionnaire_stats}
\setlength{\tabcolsep}{3pt}
\renewcommand{\tabularxcolumn}[1]{m{#1}}
\begin{tabularx}{\columnwidth}{@{} >{\raggedright\arraybackslash}X c c c c @{}}
  \toprule
  \textbf{Questions} & \textbf{FC} & \textbf{CL} & \textbf{\textit{p}} & $d_z$ \\
  \midrule
    \textbf{Q1) Preference Modeling: \newline \small Solver modelled my preferences accurately.} & $3.33{\pm}0.77$ & $\mathbf{3.65{\pm}0.71}$ & \textbf{0.013} & \textbf{0.32} \\
  \addlinespace

  \textbf{Q2) Affective Awareness: \newline \small Solver considered my emotional state.} & $3.09{\pm}1.06$ & $\mathbf{3.44{\pm}0.88}$ & \textbf{0.015} & \textbf{0.30} \\
  \addlinespace

  \textbf{Q3) Persuasiveness: \newline  \small Solver was convincing during the negotiation.} & $3.32{\pm}0.93$ & $\mathbf{3.65{\pm}0.79}$ & \textbf{0.043} & \textbf{0.26} \\
  \addlinespace

  Q4) Participant Decisions: \newline \small Solver's gestures affected my offer behavior. & $3.29{\pm}1.30$ & $3.50{\pm}1.23$ & 0.058 & $0.21$ \\
  \addlinespace

  Q5) Time Awareness: \newline \small While Solver was making its offers, it considered the time. & $3.06{\pm}1.23$ & $3.32{\pm}1.23$ & 0.121 & $0.19$ \\
  \bottomrule
\end{tabularx}
\end{table}

The unadjusted item comparisons favored CL for \textbf{Preference Modeling} ($p=0.013$, $d_z=0.32$), \textbf{Affective Awareness} ($p=0.015$, $d_z=0.30$), and \textbf{Persuasiveness} ($p=0.043$, $d_z=0.26$). Responsiveness to affective feedback may improve participants' broader impression of the agent. In contrast, \textit{Time Awareness} showed no significant difference ($p=0.121$).

\subsection{Exploratory Mediation Analysis}

To explore whether increased concessions help explain the agent's utility gain, we conducted a within-participant mediation analysis following the Montoya \& Hayes \cite{montoya2017} framework. Path $a$ was the mean within-person CL-FC mediator change; path $b$ was its coefficient in a regression of outcome change, controlling for each participant's mean mediator level ($\bar{M}_i$) and condition order. We estimated the indirect effect $ab$ and unadjusted 95\% bias-corrected and accelerated (BCa) intervals using 5,000 bootstrap resamples of whole participants, retaining both conditions (seed 42), with leave-one-participant-out jackknife acceleration.

We observed a significant total effect on Agent Utility ($c=0.022$, $p=0.017$) alongside a significantly increased concession rate ($p=0.013$, $d_z=0.31$). The tested concession-to-utility indirect effect was $ab=0.0001$ (95\% BCa CI $[-0.0067,0.0063]$), leaving the link between these gains unresolved. The 14 exploratory models paired concession rate with four outcomes (agent utility, participant utility, Nash distance, and agreement rounds), and estimated arousal with those four outcomes and the six participant move rates. Thirteen indirect-effect intervals included zero. The only interval excluding zero was for the arousal-to-Selfish-rate path ($ab=0.0862$, 95\% BCa CI $[0.0228,0.1705]$); its positive indirect effect opposes the negative total Selfish-rate difference, so it does not explain that overall decrease. This unadjusted result remains exploratory.
\section{Discussion}
\label{sec:discussion}

Our study shows the functional value of coupling continual affective personalization with differential feedback during live human-agent negotiation. Under the same negotiation policy, the CL-enabled system achieved higher agent utility and elicited higher participant concession rates than the fixed-perception baseline. Our contribution couples participant-specific model adaptation with a continuous control rule that turns angular and radial changes in valence-arousal estimates into target-utility adjustments. This extends the Solver strategy's categorical affective feedback~\cite{keskin2025} with a continuous, participant-adapted signal used throughout the interaction.

\textbf{RQ1 (Affective Feedback):} Generic emotion models face challenges in real-life, situated interactions, as highlighted by recent ``Critical Lessons'' in affective computing \cite{gunes2023}. Our study brings continual adaptation into this setting. CL yielded higher mean arousal estimates ($d_z=0.68$), with higher estimates for approximately 76\% of participants and a wider output range. It also produced positive round-level arousal-valence coupling ($r=0.45$), whereas the generic baseline showed a negative association ($r=-0.15$). Through differential feedback, these adapting estimates influence the Solver agent's target utility throughout negotiation. Changes in valence and arousal set the direction and strength of the affective feedback, giving perceptual adaptation a strategic role in offer selection.

The larger condition difference in arousal than valence motivates dimension-specific adaptation. Work on human affective responses suggests different expressive cues for these dimensions \cite{ekman1993, lang1993, scherer2005}. Participant-specific adaptation may be particularly useful for subtle arousal-related cues. This provides a rationale for concentrating updates on arousal while retaining valence information in the two-dimensional feedback rule. Future affective computing designs could test whether such selective adaptation preserves the negotiation gains while decreasing computational cost.

\textbf{RQ2 (Behavioral Influence):} The CL-enabled Solver agent used a continuously adapted affective signal to regulate its strategic stance. The control rule favors maintaining target utility when differential feedback is positive and concession when it is negative, with the affective term weighted by the opponent's behavioral responsiveness. The strategy can therefore adjust its degree of insistence as affective and offer feedback evolve. Participants increased their concession rates from 28.1\% to 37.4\%. The EASI framework links emotional displays to affective reactions and inferences about a counterpart's intentions~\cite{vanKleef2009}, providing a basis to understand how participants may respond to the agent's reactions.

A parallel EASI account connects these social responses with the agent's strategic adaptation. The mediation analysis~\cite{montoya2017} did not establish the tested concession-to-utility link, nor did it measure the EASI routes separately. The agent uses affective feedback to adjust target utility, while participants may infer its bargaining limits from its reactions and revise their offers, consistent with EASI's inferential route. Those adjustments might unfold concurrently: the control rule governs the agent's bidding, while affective reactions and inferences about the counterpart can shape participants' offers. This provides a logical explanation for utility and concession gains without requiring a serial chain linking them.

\textbf{RQ3 (Negotiation Efficiency):} The CL-enabled agent secured significantly more individual utility ($d_z=0.30$). Participant bootstrap and order-balanced estimates supported the paired utility gain. Negotiations took fewer rounds on average, although this difference was not significant ($p=0.134$). Participant utility and Nash distance also did not differ significantly. Under the shared bidding and acceptance rule, the utility gain demonstrates the value of coupling participant-adapted affective input to negotiation control. In this architecture, perception enters the decisions that determine the agent's bargaining stance throughout the exchange. This finding motivates evaluating personalization through an embodied agent's decisions and outcomes, alongside its affective signals.

\textbf{RQ4 (Questionnaire Analysis):} The CL-enabled Solver agent also changed how participants judged the interaction. Preference modeling, affective awareness, and persuasiveness received higher ratings in the exploratory item comparisons. The opponent-modeling algorithm was unchanged, yet participants rated preference modeling more highly. This pattern is consistent with a ``Halo Effect'': responsiveness to personalized affective feedback may influence judgments of broader strategic competence. This impression helps explain the higher persuasiveness ratings and offers a possible social context for the increase in concession rates.
\section{Conclusion}
\label{sec:conclusion}

This paper introduces a differential affective feedback strategy for human-agent negotiation. The strategy combines directional and intensity changes in continually personalized valence-arousal estimates to regulate target utility, with asynchronous model updates during live interaction. In a counterbalanced within-subject study ($N=66$), the CL-enabled agent achieved higher individual utility ($p=0.017$, $d_z=0.30$), and participants increased their concession rates by 9.35 percentage points ($p=0.013$, $d_z=0.31$). Both comparisons remained significant in the four-outcome Holm family. Participant utility and Nash distance did not differ significantly. CL also yielded higher arousal estimates ($d_z=0.68$; 76\% of participants) and positive round-level valence-arousal coupling ($r=0.45$). Exploratory ratings favored CL for affective awareness ($p=0.015$) and preference modeling ($p=0.013$). These findings show the value of a negotiation strategy using participant-adapted affect as an ongoing control input. The approach gives continual learning in affective robotics a strategic role in automated negotiation, with demonstrated gains in live human-robot interaction.

\subsection{Limitations \& Future Work}

This paper provides a starting point for emotionally adaptive, self-interested agents whose perceptions change with interaction. We aim to extend the CL framework to audio and text-based modalities to address current unimodal limitations. These signals would let the agent incorporate complementary cues when facial and vocal feedback differ. For example, a multimodal model could process a smiling face paired with an angry vocal tone, directly tackling recent calls for multimodal robotics \cite{churamani2022}. We also plan longitudinal studies to test whether personalized models persist across ongoing buyer-seller relationships, where adaptation can support repeated interaction.

We selected FaceChannel as both the baseline and pseudo-label source for two reasons: its continuous valence-arousal regression fits the circumplex design, and its lightweight CNN allows real-time, on-device inference, preserving privacy while meeting our $<2.0$\,s latency constraint. Our differential-feedback rule operates on valence-arousal outputs, so it can be paired with other perception backbones that provide these estimates. Here, CLIFER supplies episodic adaptation and generative replay. Evaluating alternative backbones, including modern foundation models, would test how the same control principle transfers across perceptual representations.

Episodic Memory smooths frame-level pseudo-labels via competitive clustering, but its representational capability is limited by the frozen encoder. If the generic base model shows zero-shot failure for a participant, clustering cannot create an affective signal that is absent from the latent space. Our evaluation focused on functional coherence and negotiation utility without an independent affect criterion; output range and correlation therefore do not establish recognition precision. Personalization concerns the current participant and interaction, with cross-session and cross-context persistence remaining questions for longitudinal evaluation. Retention and catastrophic forgetting were not separately evaluated.

Our approach uses a brief demo session to train CLIFER on each participant before the main negotiation, addressing the cold-start problem. While such a demo session is feasible in a controlled lab setting, real-life scenarios commonly lack a warm-up interaction. The comparison evaluates demo-session training and subsequent online adaptation together. Comparing demo-trained CLIFER held fixed during negotiation with the same model allowed to adapt would separate the contribution of subsequent updates. Comparing the shared differential controller with categorical or non-affective feedback would separately test the control rule’s contribution. To reduce the need for a demo session, future work should pursue few-shot meta-learning (e.g., MAML~\cite{finn2017}) to support deployment with fewer initial observations.

Our claims of external validity are limited to short-term, lab-based negotiations with university students from one country and a specific robotic embodiment. Moreover, cultural and educational norms influence conflict resolution \cite{gal2011, haim2012}. For example, a smile may show agreement within one culture but deception in another; thus, cross-cultural validation is needed to determine whether the CL mechanism can adapt to expressive differences. Combining the feedback rule with language-based argument generation \cite{dogru2025} could extend adaptation from offer selection to verbal negotiation. Using the personalized emotion score ($P_E$), the agent could adjust its explanations alongside its bids, for example by responding to changes in estimated affect with calming language \cite{qian2025}.

The closed loop is central to the proposed approach: affective estimates influence offers, which can alter participants' behavior and subsequent facial input. We therefore evaluate the complete CL-enabled negotiation system under a shared control rule. Counterbalancing distributes common session effects across conditions; FC-session facial expressions were excluded from CL training. Condition-specific human carryover remains possible.

\subsection{Ethical Impact \& Dual-Use}

This work presents an agent that uses facial affect estimates to gain a competitive advantage.
The utility gains demonstrate the strategic value of personalized affective feedback, but the technology also has dual-use potential. A fully self-interested CL-enabled Solver Agent could exploit signs of mental fatigue or frustration, pushing aggressive offers when the human partner is vulnerable. To limit this risk, real-world deployments must follow principles of algorithmic accountability.
Systems should obtain informed consent and clearly notify users that they use real-time estimates from facial expressions to adapt the negotiation strategy.

\section*{Acknowledgment} 
\noindent\textbf{Funding:} This work has been supported by the Scientific and Technological Research Council of T\"urkiye (T\"UB\.{I}TAK) with grant number 118E197. N. Churamani and H. Gunes were supported in part by the EPSRC project ARoEQ under grant ref. EP/R030782/1. C. M. Jonker's contribution to this research was partially supported by NWO under Grant Numbers 024.004.022 (Hybrid Intelligence), 024.004.031 (ESDiT), and 024.005.017 (ALGOSOC). The support is gratefully acknowledged. 

\noindent\textbf{Open Access:} For open access purposes, the authors have applied a Creative Commons Attribution (CC BY) licence to any Author Accepted Manuscript version arising.

\noindent\textbf{Data Access:} Source code, analysis scripts, and anonymized estimated valence-arousal scores and negotiation records from 66 participants are available at \url{https://github.com/monurkeskin/Personalized-Affective-Perception-via-Continual-Learning-TAFFC-2026}. We processed participants' facial images in memory and did not share them in the repository, in accordance with the study's ethics approval and participant consent.

\noindent\textbf{AI Usage:} We acknowledge the use of AI for grammatical editing, proofreading, and formatting optimization of the manuscript; all intellectual contributions, experimental designs, data analyses, and scientific conclusions were conceived, executed, and verified only by the human authors. Any opinions, findings, and conclusions, or recommendations expressed in this material are those of the author(s) and do not necessarily reflect the views of the supporting organizations.

\bibliographystyle{IEEEtran}
\bibliography{References}

\vspace{-24pt}
\begin{IEEEbiography}[{\includegraphics[width=1in,height=1.25in,keepaspectratio]{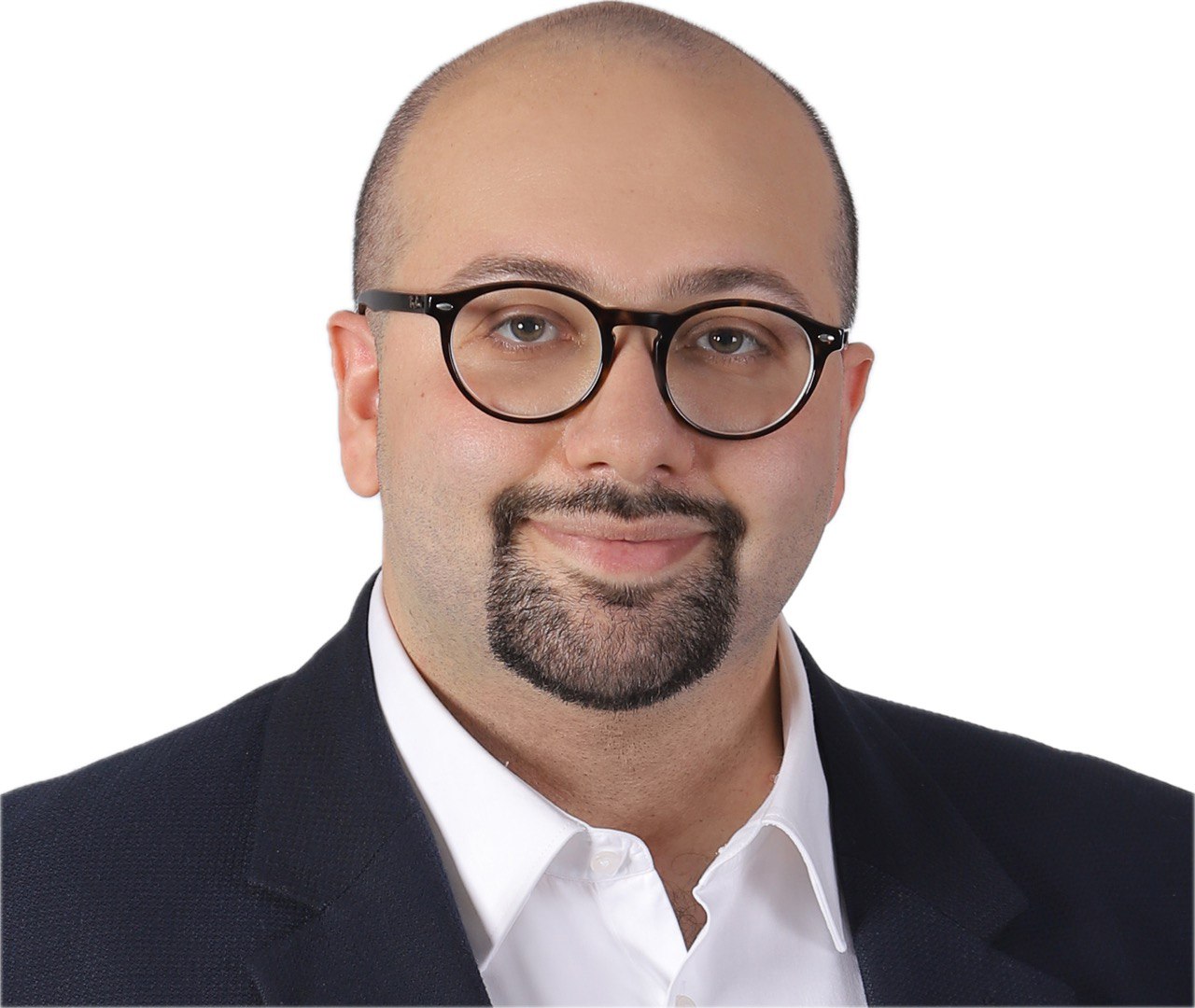}}]{Mehmet Onur Keskin} received the B.Sc. degree in computer science in 2020 and the M.Sc. degree in artificial intelligence in 2023 from \"Ozye\u{g}in University, Istanbul, T\"{u}rkiye. He worked as a research assistant in its Interactive Intelligent Systems Laboratory and held visiting research positions at the University of Cambridge, TU Delft, Brown University, and Oregon State University. His interests include multi-agent systems, human-robot cooperation, and decision-making under uncertainty. He received the IEEE Computer Society Richard E. Merwin Student Scholarship and \"Ozye\u{g}in University's Best Research Assistant Award in 2022. He organized the Automated Negotiating Agents Competition at AAMAS 2023 and was selected as a young researcher for the 2025 Heidelberg Laureate Forum.
\end{IEEEbiography}

\vspace{-24pt}
\begin{IEEEbiography}[{\includegraphics[width=1in,height=1.25in,keepaspectratio]{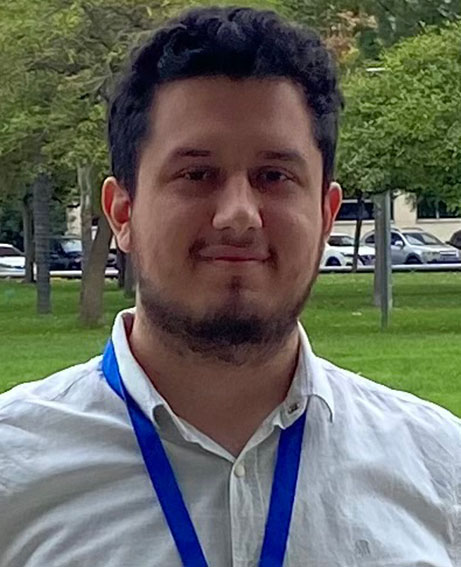}}]{Berk Buzcu}
earned his Bachelor of Science in Computer Science from \"Ozye\u{g}in University. He is a doctoral researcher at HES-SO Valais-Wallis, Switzerland, and an external Ph.D. candidate at Technische Universit\"at Berlin, Germany. During his master's studies at \"Ozye\u{g}in University, he participated in the CHIST-ERA (locally T\"UB\.{I}TAK) funded project, ``EXPECTATION: Personalized Explainable AI for decentralized agents with heterogeneous knowledge.'' Within the project, he focused on applying explainability techniques in recommender systems, in line with his research interests. The main goal was to enhance the success of recommender systems in achieving user goals via sociability.
\end{IEEEbiography}

\vspace{-24pt}
\begin{IEEEbiography}[{\includegraphics[width=1in,height=1.25in,keepaspectratio]{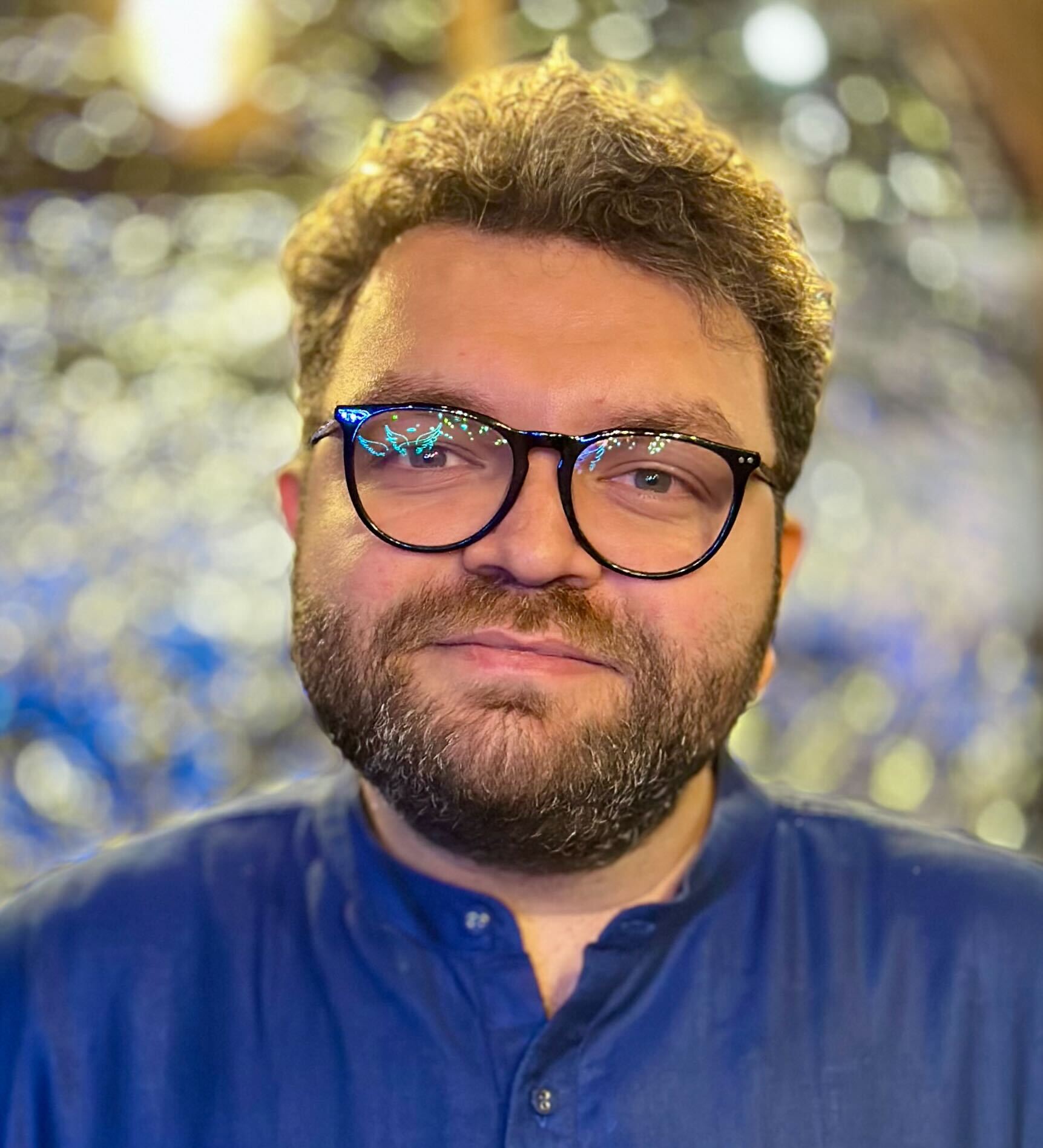}}]{Nikhil Churamani} is a Visiting Researcher at the Affective Intelligence and Robotics (AFAR) Laboratory, University of Cambridge. His research interests include Computer Vision, Affective Robotics, Continual Lifelong Learning, Federated Learning, Human-in-the-loop Reinforcement Learning and Human-Robot Interaction. Previously, he was a Postdoctoral Researcher at the AFAR lab at the University of Cambridge, supported by Google under the GIG funding scheme, exploring Federated Continual Learning of socially appropriate robot behaviors in human-centered environments. His PhD research at the University of Cambridge focused on Continual Learning for Affective Robotics, funded by the Engineering and Physical Sciences Research Council, UKRI.
\end{IEEEbiography}

\vspace{-24pt}
\begin{IEEEbiography}[{\includegraphics[width=1in,height=1.25in,keepaspectratio]{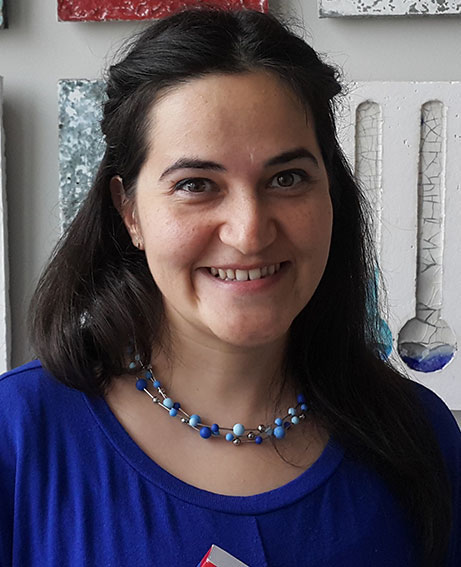}}]{Associate Professor Reyhan Aydo\u{g}an} is a faculty member and chair of the Department of Artificial Intelligence and Data Engineering at \"Ozye\u{g}in University in \.{I}stanbul, T\"urkiye. She is also a research fellow in the Interactive Intelligence Group at Delft University of Technology. Dr. Aydo\u{g}an received her doctorate in Computer Engineering from Bo\u{g}azi\c{c}i University in 2011. Following her Ph.D., she worked as a postdoctoral researcher on various AI projects at TU Delft. Her research interests include explainable interactive intelligent systems, decision support systems, agreement technologies (with a particular focus on negotiation), and human-agent interaction. Dr. Aydo\u{g}an is an active member of the program committees of leading AI conferences, including AAAI, IJCAI, AAMAS, ECAI, ACII, and PRIMA. She is also one of the main organizers of the International Automated Negotiating Agents Competition (ANAC).
\end{IEEEbiography}

\vspace{-24pt}
\begin{IEEEbiography}[{\includegraphics[width=1in,height=1.25in,keepaspectratio]{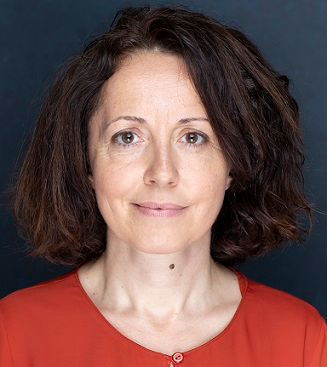}}]{Hatice Gunes} (Senior Member, IEEE) is a Full Professor of Affective Intelligence and Robotics, Director of \href{https://cambridge-afar.github.io/} {Cambridge AFAR Lab} and Associate Director of the Centre for Human-Inspired Artificial Intelligence (CHIA) at the University of Cambridge, UK. She previously served as President of the Association for the Advancement of Affective Computing, as a Faculty Fellow of the Alan Turing Institute and a Fellow of the EPSRC. Her research advances multimodal, social and affective intelligence for AI systems and robots, with a strong emphasis on wellbeing, ethics, fairness and explainability, and has received many honours and awards, including being named an Honoree for the Sony Women in Technology Award with Nature 2025.
\end{IEEEbiography}

\vspace{-24pt}
\begin{IEEEbiography}[{\includegraphics[width=1in,height=1.25in,keepaspectratio]{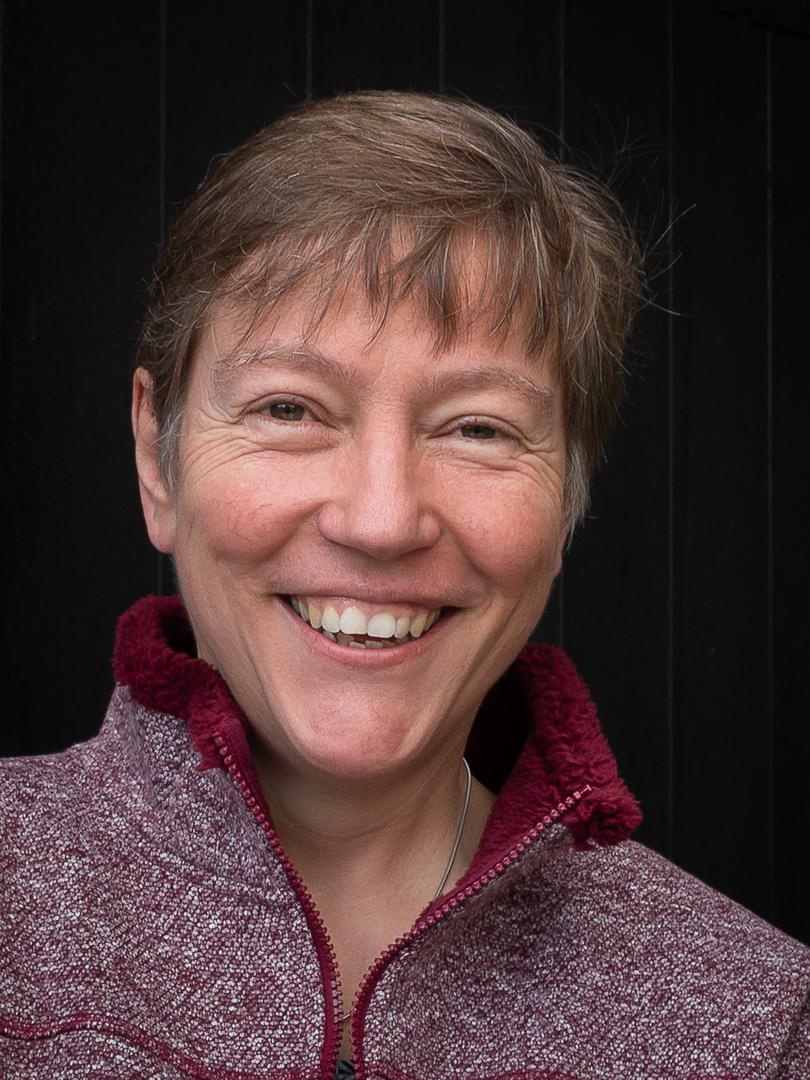}}]{Catholijn M. Jonker} is a Professor of Interactive Intelligence at TU Delft, Explainable Artificial Intelligence at Leiden University, and Hybrid Intelligence at Vrije Universiteit Amsterdam. She received her Ph.D. degree from Utrecht University. Her research focuses on decision and negotiation support systems, human-agent teamwork, and hybrid intelligence. She coordinates the Hybrid Intelligence Center and is a principal investigator at the Amsterdam Institute for Advanced Metropolitan Solutions. She is a Fellow of the European Association for Artificial Intelligence and a member of Academia Europaea. She received the ACM Autonomous Agents Research Award in 2024.
\end{IEEEbiography}

\end{document}